\documentclass[8.5pt,twoside]{article}

\usepackage[super,sort&compress,comma]{natbib} 
\usepackage{times,mathptm}
\usepackage{sectsty}
\usepackage{balance} 
\usepackage[text={11.3cm,20.4cm},centering]{geometry} 
\usepackage[T1]{fontenc}

\usepackage{graphicx} 
\usepackage{lastpage}
\usepackage[format=plain,justification=raggedright,singlelinecheck=false,font=small,labelfont=bf,labelsep=space]{caption} 
\usepackage{fancyhdr}
\usepackage[colorinlistoftodos]{todonotes}

\usepackage[colorlinks=true,linkcolor=blue,citecolor=blue]{hyperref}

\begin{document}

\thispagestyle{plain}
\fancypagestyle{plain}{
\fancyhead[L]{\includegraphics[height=8pt]{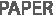}}
\fancyhead[C]{\hspace{-1cm}\includegraphics[height=15pt]{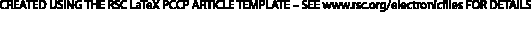}}
\fancyhead[R]{\includegraphics[height=10pt]{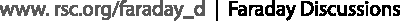}\vspace{-0.2cm}}
\renewcommand{\headrulewidth}{1pt}}
\renewcommand{\thefootnote}{\fnsymbol{footnote}}
\renewcommand\footnoterule{\vspace*{1pt}%
\hrule width 11.3cm height 0.4pt \vspace*{5pt}} 
\setcounter{secnumdepth}{5}

\makeatletter 
\renewcommand{\fnum@figure}{\textbf{Fig.~\thefigure~~}}
\def\subsubsection{\@startsection{subsubsection}{3}{10pt}{-1.25ex plus -1ex minus -.1ex}{0ex plus 0ex}{\normalsize\bf}} 
\def\paragraph{\@startsection{paragraph}{4}{10pt}{-1.25ex plus -1ex minus -.1ex}{0ex plus 0ex}{\normalsize\textit}} 
\renewcommand\@biblabel[1]{#1}            
\renewcommand\@makefntext[1]%
{\noindent\makebox[0pt][r]{\@thefnmark\,}#1}
\makeatother 
\sectionfont{\large}
\subsectionfont{\normalsize} 

\fancyfoot{}
\fancyfoot[LO,RE]{\vspace{-7pt}\includegraphics[height=8pt]{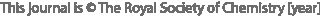}}
\fancyfoot[CO]{\vspace{-7pt}\hspace{5.9cm}\includegraphics[height=7pt]{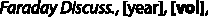}}
\fancyfoot[CE]{\vspace{-6.6pt}\hspace{-7.2cm}\includegraphics[height=7pt]{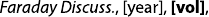}}
\fancyfoot[RO]{\scriptsize{\sffamily{1--\pageref{LastPage} ~\textbar  \hspace{2pt}\thepage}}}
\fancyfoot[LE]{\scriptsize{\sffamily{\thepage~\textbar\hspace{3.3cm} 1--\pageref{LastPage}}}}
\fancyhead{}
\renewcommand{\headrulewidth}{1pt} 
\renewcommand{\footrulewidth}{1pt}
\setlength{\arrayrulewidth}{1pt}
\setlength{\columnsep}{6.5mm}
\setlength\bibsep{1pt}

\noindent\LARGE{\textbf{How to do impactful research in artificial intelligence for chemistry and materials science}}

\vspace{0.6cm}
\noindent\large{\textbf{Austin H. Cheng,\textit{$^{a,b,c}$}
Cher Tian Ser,\textit{$^{a,b,c}$}
Marta Skreta,\textit{$^{b,c}$} \\
Andrés Guzmán-Cordero,\textit{$^{c,d}$}
Luca Thiede,\textit{$^{b,c}$}
Andreas Burger,\textit{$^{b,c}$} \\
Abdulrahman Aldossary,\textit{$^{a}$}
Shi Xuan Leong,\textit{$^{a,e}$} \\
Sergio Pablo-García,\textit{$^{f}$}
Felix Strieth-Kalthoff,\textit{$^{g}$} \\
and Alán Aspuru-Guzik\textit{$^{a,b,c,f,h}$}}}\vspace{0.5cm}

\noindent\textit{\small{\textbf{Received Xth XXXXXXXXXX 20XX, Accepted Xth XXXXXXXXX 20XX\newline
First published on the web Xth XXXXXXXXXX 200X}}}

\noindent \textbf{\small{DOI: 10.1039/c000000x}}
\vspace{0.6cm}

\noindent \normalsize{
Machine learning has been pervasively touching many fields of science. Chemistry and materials science are no exception.
While machine learning has been making a great impact, it is still not reaching its full potential or maturity. 
In this perspective, we first outline current applications across a diversity of problems in chemistry. Then, we discuss how machine learning researchers view and approach problems in the field.
Finally, we provide our considerations for maximizing impact when researching machine learning for chemistry.
}
\vspace{0.5cm}

\footnotetext{\textit{$^{a}$~Department of Chemistry, University of Toronto, Toronto, Ontario M5S 3H6, Canada.}}
\footnotetext{\textit{$^{b}$~Department of Computer Science, University of Toronto, Toronto, Ontario M5S 2E4, Toronto, Canada.}}
\footnotetext{\textit{$^{c}$~Vector Institute for Artificial Intelligence, Toronto, Ontario M5G 1M1, Canada.}}
\footnotetext{\textit{$^{d}$~Tinbergen Institute, University of Amsterdam, Amsterdam, Netherlands.}}
\footnotetext{\textit{$^{e}$~School of Chemistry, Chemical Engineering and Biotechnology, Nanyang Technological University, Singapore 63737.}}
\footnotetext{\textit{$^{f}$~Acceleration Consortium, Toronto, Ontario M5G 1X6, Canada.}}
\footnotetext{\textit{$^{g}$~School of Mathematics and Natural Sciences, University of Wuppertal, Wuppertal, Germany.}}
\footnotetext{\textit{$^{h}$~Department of Chemical Engineering and Applied Chemistry, University of Toronto. Department of Materials Science and Engineering, University of Toronto. Lebovic Fellow, Canadian Institute for Advanced Research (CIFAR).}}

\section{Introduction}

Machine learning (ML) has been applied in many facets of chemistry, and its use is rapidly growing. We argue in this perspective that despite this dramatic growth and impact, ML could be employed better and more extensively. Current work is still far from exhausting the potential of ML to advance theory and application in chemistry in terms of breadth, depth, and scale. In addition, the actual types of problems that ML could tackle, such as hypothesis generation or enabling internalized scientific understanding, are still areas of active research or open problems.

To color a picture of the field, we begin by outlining a taxonomy of the chemical problems to which ML has been applied, ranging from prediction, generation, synthesis, force fields, spectroscopy, reaction optimization, and foundation models.
Shifting gears, we then introduce types of problems in ML and show how chemical problems can be reformulated as instances of ML problems.
These standard problems help organize the toolbox of algorithms and theory provided by ML.
Digging further into this perspective, we examine differences in practices and values between the ML and chemistry communities and highlight where collaboration and cross-pollinating perspectives can advance both fields.
Armed with the above, we can then discuss how to select impactful applications of ML in chemistry and recommend our suggested good practices for research in this area.

\section{Chemistry meets data: A taxonomy of problems}
\label{SEC:taxonomy}
\begin{figure}[h]
\centering
\makebox[\textwidth][c]{
  \includegraphics[width=1.25\textwidth]{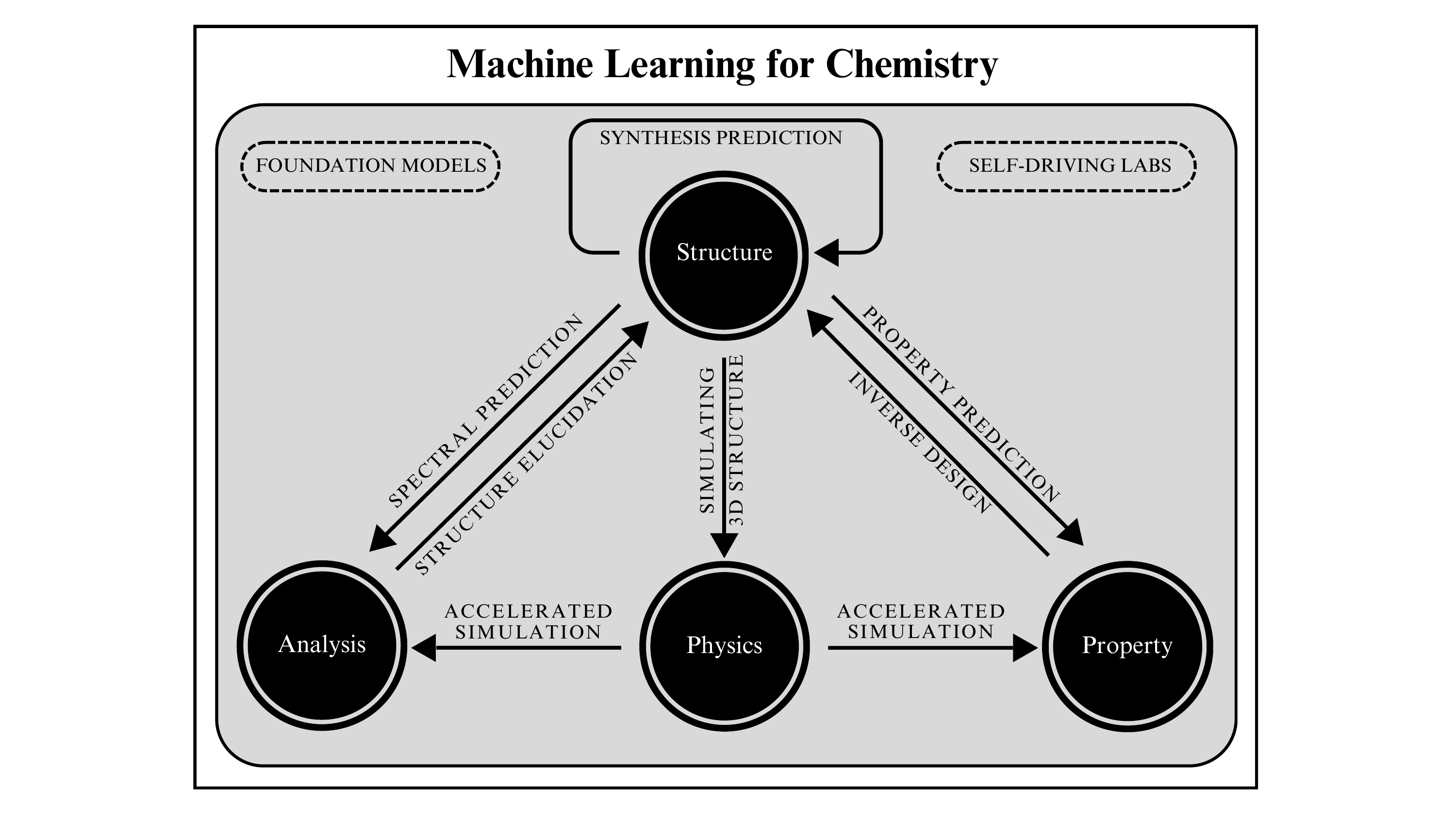}
  }
  \caption{A taxonomy of chemical problems related to machine learning. Each arrow indicates an application of ML and signifies how all these relate to each other. Foundation models and self-driving labs touch all these areas.}
  \label{fgr:taxonomy}
\end{figure}

Chemistry, and science in general, involves data in one form or another. Not surprisingly, then, data science is integral to chemistry. Machine learning, a subfield of data science, has become an integral tool in our domain science's arsenal. Therefore, it is crucial to begin cataloguing and organizing critical efforts to date.

We suggest a taxonomy of the chemical problems to which machine learning has been applied.
As shown in Figure \ref{fgr:taxonomy}, ML has been applied to solve various chemical problems by encoding and decoding to and from chemical structure, properties, 3D structure and dynamics, and experimental data.
For reasons of space, time, and focus, this is not a comprehensive review but rather an opportunity to highlight diverse applications of ML in chemistry.
We will not introduce ML algorithms in detail. 
For exhaustive reviews, please see other works.\cite{coley2020autonomous1, coley2020autonomous2, wang2023scientific, aldossary2024silico, strieth2020machine, tom2024self}


\subsection{Structure to property: property prediction}

\subsubsection{Cheminformatics and quantitative structure-activity relationships.~~}

Chemistry has leveraged data to predict properties from a chemical structure long before the everyday use of the term ``machine learning''. This field has been originally identified initially as \emph{cheminformatics}. 
These tools sought to store, retrieve, and model chemical structures.
Early examples began in 1957 with substructure searches in a database,\cite{ray1957finding} followed by simple multivariate regression for learning quantitative structure-activity relationships\cite{kubinyi2002narcosis} (QSAR) between molecular descriptors like Hammett constants and partition coefficients, and biological activity.\cite{hansch1962correlation, hansch1964p} These were mostly property-activity relationships -- the first structure-activity relationships involved local explanations analyzing how substituents on a ring affected activity,\cite{free1964mathematical} which could be generalized to many scaffolds via substructural analysis.\cite{cramer1974substructural}
Eventually, computers automatically encoded molecular structures as fingerprints -- bit-vectors that store the presence or absence of many substructures found in the molecule.\cite{rogers2010extended}
These fingerprints were useful in encoding molecular structures to predict molecular activity in simple models such as support vector machines.\cite{glick2006enrichment}

\subsubsection{Representing molecules with expert descriptors.~~}

While chemists have a conceptual understanding of the effects of functional groups on the properties of a molecule, communicating this information to a model is critical to ensure that the model is predictive. 
Expert descriptors infuse chemical knowledge derived from experiments or conceptual knowledge into the features provided to a model and have achieved good predictive performance, especially in low-data regimes. These expert descriptors also generalize well outside the model's training set, as chemical knowledge is baked into these features. As early as 1937, Hammett fitted sigma parameters for predicting the influence of chemical substituents on reactivity.\cite{hammett1937effect} Additionally, group contribution methods, which assume that structural components or functionalizations behave the same way across many different molecules, parameterize these components into numerical features that can be used to predict molecular properties.\cite{bruice1956correlation, ambrose1978correlation, nannoolal2004estimation} The discipline has since grown to involve molecular fingerprinting techniques and the incorporation of 2D and even 3D information for use in prediction. In more recent times, as the properties of a homogeneous transition metal catalyst are strongly influenced by the ligands attached to it, parameterizing the structural and electronic features of these phosphine ligands has also been successful in predicting the properties of a catalyst.\cite{gensch2022comprehensive, tolman1970electron, tolman1970phosphorus} 
Looping back to historical models, recent work has also been able to leverage density functional theory (DFT) and machine learning to successfully machine learn Hammett parameters.\cite{monteiro2023machine}

\subsubsection{Learned chemical representations.~~}

Models have become more complex with advances in computational hardware, moving from simple linear regression models to complex architectures like auto-encoders, generative adversarial networks, graph neural networks or transformers. Instead of relying on chemists to intuit the best way to represent a molecule, we can now harness the ability of models to automatically learn and exploit complex patterns within large amounts of data for property prediction. To a certain level of abstraction, which tends to ignore 3D information or wave function properties, molecules can be naturally represented as graphs where atoms are nodes and bonds are edges. By relaxing the notion of fingerprints from discrete bit-vectors to continuous feature vectors, we proposed graph neural networks to automatically learn continuous representations of important substructures, achieving state-of-the-art performance on molecular property prediction tasks.\cite{duvenaud2015convolutional, yang2019analyzing} These representations have been deployed widely across multiple avenues like machine learning for olfactory properties of a molecule, \cite{lee2023principal} and in catalysis where adsorption properties of adsorbates were predicted. \cite{pablo2023fast}

While simple atomic and bond features required for the constructed graphs can be generated quickly,\cite{heid2023chemprop} the properties that one wants to target for prediction are much harder to obtain -- especially in higher qualities and fidelities. As learned representations typically require large amounts of data, complicated architectures do not function as well with low amounts of data gathered from typical experimental settings. To bridge this gap, molecular benchmarks were created to assess the quality of such learned representations properly. These benchmarks contain tasks gathered from literature data related to predicting biological behaviours and physicochemical or quantum chemical properties and provide a common ground on which different machine-learning architectures can harness and exploit the same data in various ways for property prediction. \cite{wu2018moleculenet} 

To improve the performance of such graph embeddings, they can be further tuned if there are some intuitions about how the embedding spaces should be reshaped to reflect the distances between inputs properly. These can involve strategies like making the embeddings aware of how chemical reactions should transform these embeddings \cite{wang2021chemical} or through strategies like contrastive learning. \cite{wang2022molecular} Finally, for tasks sensitive to the molecule's conformation in three dimensions, incorporating three-dimensional representations that exceed the capability of the innately deficient two-dimensional graphs has proven successful in predicting molecular properties. \cite{zhou2023unimol}

\subsubsection{Limits and open problems.~~}

Despite the great strides made in molecular machine learning, the ability of machine learning models to extrapolate beyond the data it is trained on is still limited, posing barriers for application to novel chemistries. Several approaches can potentially bridge these gaps. For example, by using physics-informed models that can contain fundamental representations that help in generalizing the representation itself to satisfy some symmetries or properties related to the physical laws of nature. Active learning is also a powerful tool for expanding datasets on the fly by capturing computational or experimental data for extrapolation.
Additionally, while models have progressively performed better on property prediction benchmark tasks, these benchmarks represent only a tiny subset of chemical tasks, making their performance on various other tasks unknown.\cite{practicalcheminformaticsNeedBetter} While we have attempted to create benchmarks more representative of typical tasks,\cite{nigam2023tartarus} this is still not a central focus of the community.

Structure-to-property models have been widely employed in screening projects, leading to experimentally verified predictions. We will discuss a few selected case studies in Sec. \ref{SEC:Screening}. 

\subsection{Property to structure: designing molecules in chemical space}

While the rational design paradigm analyzes the relationship between structure and properties to design promising molecules, another paradigm asks: what are all the possible molecules that satisfy a given property? Solving this question is known as the {\sl inverse design} problem.

Chemical space is the set of all synthesizable molecules and is often cited for having an astronomical size of at least $10^{33}$ to $10^{60}$ molecules.\cite{polishchuk2013estimation,bohacek1996art}
Within this vast space are potential drugs that could cure current diseases and putative materials that could enable a sustainable future.


\subsubsection{Virtual screening.~~}
\label{SEC:Screening}
A simple approach to navigating chemical space is to enumerate a feasible set of possible options and then narrow them down to the best solution.
This shift in perspective has its experimental implementation employing strategies such as high-throughput screening of chemical libraries and combinatorial chemistry to synthesize these libraries.\cite{warr1997combinatorial}
Given the astronomical size of chemical space, it became clear that arbitrarily searching through compounds would produce few promising hits, making this approach inefficient as the cost of extensive chemical synthesis campaigns is often taxing or prohibitive.\cite{carroll2005will}
This motivated virtual screening and computational search funnels as a way to filter out unpromising compounds, leaving only the best candidates for synthesis and testing.
In drug discovery, molecules are filtered out with computationally lean checks such as high molecular weight or problematic functional groups, followed by more computationally intensive docking for estimating binding affinity, ultimately narrowing down to a handful of lead compounds.\cite{walters1998virtual}
Scaling the size of virtual libraries increases the likelihood of promising hits, which has motivated ever-larger screening campaigns requiring increasing computational resources.
One example was the Harvard Clean Energy Project,\cite{hachmann2011harvard} in which we searched through $10^7$ candidates with quantum chemistry calculations on distributed volunteer computing to search for efficient organic photovoltaics.

Similarly, VirtualFlow\cite{gorgulla2020open} docked over $10^9$ molecules by efficiently using thousands of CPU cores.
As the size of chemical libraries grows, with the required computational resources scaling linearly, hierarchical approaches to evaluate the fitness of individual synthetic building blocks offer a way past linear scaling.\cite{sadybekov2022synthon}

\subsubsection{Generative models for inverse design.~~}

As the size of chemical libraries surpasses $10^{15}$ molecules\cite{sadybekov2023computational} and becomes computationally prohibitive to screen, ML offers ways to consider large search spaces without simulating all molecules.
For example, in a chemical library, many molecules should have similar structures and properties, so running simulations on every molecule is redundant.
A formal way to handle this is to simulate a portion of the library and then train property prediction models on this subset, which should be generalized across the library. Since these property prediction models are computationally cheaper than simulations, they can be evaluated for the entire library and used to prioritize candidates for simulation.
We leveraged this approach to design organic light-emitting diodes that were verified experimentally.\cite{gomez2016design}

However, another arm of ML offers a way to consider {\sl all} (or a vast subset) of the chemical space.
Given a dataset of molecules in a representation such as SMILES strings, generative models learn to generate strings which resemble the dataset.
Because generative models can consider arbitrary strings, they could potentially generate any molecule in chemical space.
They can also be conditioned to generate molecules with desired properties -- essentially reversing the property prediction process.\cite{sanchez2018inverse,zunger2018inverse}
Molecular generative models have been applied with many model classes. We pioneered the use of  variational autoencoders (VAEs)\cite{gomez2018automatic} for this purpose. Other examples include autoregressive models,\cite{segler2018generating} generative-adversarial networks (GANs),\cite{sanchez2017optimizing} and reinforcement learning,\cite{olivecrona2017molecular,blaschke2020reinvent} amongst many other sampling strategies. Generative models have also been extended and shown to work well with various representations like SMILES, SELFIES,\cite{krenn2020self} and Group SELFIES\cite{cheng2023group} strings, as well as molecular graphs and fragments.
Molecular optimization methods such as genetic algorithms\cite{jensen2019graph} and Bayesian optimization\cite{korovina2020chembo} also have been sometimes called generative models despite not learning a molecular distribution {\sl per se}.
A recent review of different generative model classes and representations can be found in Gao \textit{et al.},\cite{gao2022sample} although this is a rapidly moving field.

As more generative models were proposed, benchmarks such as GuacaMol \cite{brown2019guacamol} and MOSES\cite{polykovskiy2020molecular} began evaluating and comparing different generative models based on validity, novelty, uniqueness, and goal-directed optimization. Optimization has been such a primary focus that molecular design can be regarded as a combinatorial optimization of molecular properties over the space of molecular graphs.
In this way, a new benchmark emphasizes sample efficiency, which is the number of property evaluations needed to reach optimal molecules.\cite{gao2022sample}
In addition, more realistic benchmark tasks relying on simulation have been recently proposed by us in the Tartarus benchmark set.\cite{nigam2023tartarus}
Tartarus more closely resembles real-world scenarios where computational and experimental resources are constrained.

However, by departing from chemical libraries for the entire chemical space, generative models relaxed the crucial constraint of synthesizability.
Generative models can suggest molecules which are difficult to synthesize and evaluate.\cite{gao2020synthesizability}
To overcome this, synthesizable generative models consider chemical synthesis pathways when generating molecules, ensuring that the generated molecules are not only theoretically valid but also practicably synthesizable.\cite{bradshaw2019model, gao2021amortized, koziarski2024rgfn}
Other approaches combine virtual libraries with generative approaches to ensure that proposed molecules are always from the library.\cite{pedawi2022efficient} These methods have particular relevance for high-throughput arrays and self-driving laboratories, as predicted molecules that are not synthetically feasible with readily available platforms could slow down closed-loop approaches.

For a comprehensive overview of these advancements and the state of the art in molecule design, Du {et al.} provide an excellent review, summarizing the latest developments and methodologies in the field. \cite{du2024machine}

Generative models have proven worthy in the recent years. Quite notably the company InSilico Medicine has employed them to generate several drugs that are undergoing clinical trials currently. In 2019, together with InSilico and Wuxi Apptec researchers, we showed the ability of generative models to develop a lead drug candidate in approximately 45 days.\cite{zhavoronkov2019deep} Many researchers since then have continued to show other examples of generative models in drug discovery. For example, Barzilay and co-workers have developed antibiotics using similar approaches.\cite{stokes2020deep}

\subsubsection{Limits and open problems.~~}

While candidates can be generated easily with such models, the quality of the candidates depends on the ability to develop a properly performing and scalable cost function for conditioning the generative models. Additionally, these models are trained on approximate metrics, which means that their real-life performance still has to be evaluated. Thus, evaluating the synthesizability of a candidate or providing steps to make candidates is of paramount importance (see next section).

Most generative models have been developed with simple benchmarks in mind, such as predicting simple properties like logP. However, developing using proper benchmarks (such as Tartarus) or restricting them to feasible sets of molecules, such as those synthesizable with self-driving labs (see Sec. \ref{sec:sdl}), remains a challenge.

\subsection{Structure to structure: synthesis planning and reaction condition prediction}

Synthesis planning – i.e. finding synthetic pathways that give rise to a desirable target molecule – is an open challenge that chemists have faced for over a century, particularly in the ``molecular world'' of drug discovery, agricultural chemistry or molecular materials chemistry. 
This problem is complex in two respects: 
First, predicting the outcome of a specific unseen reaction, given all reactants, reagents, and reaction conditions, is effectively an unsolved problem to date.
Second, even with such a ``reaction prediction'' tool at hand, finding feasible multi-step sequences of reactions that eventually enable the synthesis of the target molecule from cheap and commercially available precursors requires searching a massive network of possible pathways. 
Additional challenges arise from practical demands to the synthesis planning problem: efficiency, cost, waste production, sustainability, safety, or toxicity are practical concerns, especially in an industrial setting.   

\subsubsection{Synthesis planning.~~} 

Synthesis planning is classically addressed through the formalism of \textit{retrosynthesis}, as pioneered by Nobel Prize winner E. J. Corey:\cite{corey1991logic} 
Using knowledge of chemical reactivity, the target molecule is gradually disconnected into progressively simpler precursors, which eventually yields commercially available starting materials. 
Formally, this corresponds to a tree search problem. 
As early as in the 1960s, Corey realized that this approach is ideally suited to be tackled in a computational manner.\cite{retrosynthesis1969}
Since then, a number of expert systems have been developed to guide this tree search.\cite{Todd2005}

The past decade has seen significant progress in addressing this challenge using the toolbox of ML. 
In this context, the key ``decision policy'' has often been treated as a multi-task regression problem: 
Given the structure of a target molecule, a ML model is trained to predict an applicable reaction out of a catalog of reactions.\cite{Wei2106, segler2017neural, coley2017prediction} 
This symbolic approach, however, requires a pre-defined catalogue of all reaction types, often referred to as reaction ``rules'' or ``templates'', which itself presents new obstacles. There is neither a generally accepted definition of the term ``reaction rule'' nor an unambiguous procedure to perform reaction rule extraction from data.
Alternatively, ``template-free'' approaches to the one-step reaction prediction problem, predict reactions as graph edits in the starting material graph,\cite{coley2019graph} or solve a sequence-to-sequence ``product-to-starting-material'' translation task.\cite{Pande, schwaller2019molecular} 
Notably, these models (template and template-free) can be similarly trained in the forward direction, predicting reaction products from starting materials. 

These single-step prediction models have been used to build tree search models, which aim to solve the full synthesis planning problem. 
In this context, a Monte-Carlo tree search is usually the method of choice. 
Following the pioneering works from Segler \textit{et al.} \cite{segler2018planning} and Coley \textit{et al.},\cite{Coley2019Robotic} a number of mostly open-source systems have been released.\cite{ Genheden2020, D0SC05078D}

\subsubsection{Prediction and optimization of reaction conditions.~~} 

What is often overlooked in synthesis planning is that knowing a possibly suitable reaction type alone does not guarantee that the envisioned intermediate or target product can be prepared from the proposed starting materials. 
The question if the product can be obtained (ideally in high yield), crucially depends on what is often referred to as the \textit{reaction conditions}: the choice of reagent(s), catalyst(s), additive(s) and solvent,  the values of continuous parameters such as stoichiometries, temperature and reaction time, as well as the practical details of running the reaction in the laboratory. 
In an ideal scenario, an AI-assisted tool would take in a new ``starting-material-to-product'' transformation, and spit out the required reaction conditions for this transformation. 
However, this is yet to be achieved, particularly because reaction conditions cover a vast combinatorial parameter space and are frequently governed by underlying physical principles that are difficult to simulate.
In practice, reaction conditions are often selected by employing ``nearest-neighbor'' reasoning based on literature precedents, either automatically or through human expertise. 

Machine learning approaches to reaction condition optimization have thus mainly focused on regression modelling of reaction yields as a function of reaction conditions. In this context, data-driven approaches have intersected with regression techniques from physical organic chemistry, which attempt to model reaction outcomes based on mechanistic considerations.
In highly constrained condition spaces, purely data-driven, supervised learning of product yields on systematically generated data from high-throughput experimentation has shown promising results.\cite{ahneman2018predicting, Granda2018, Zahrt2019, Sandfort2020, Schwaller2021Yield} For example, our work on optimizing the E/Z ratio of a reaction relevant to pharmaceutical process chemistry showed that only with $\approx 100$ experiments we were able to outperform what had been the state-of-the-art for this process by human-only reaction optimization.\cite{christensen2021data}
Meanwhile, the use of literature data for the same purpose is highly flawed,\cite{beker2022machine, anie.202204647} usually necessitating individual, case-by-case reaction optimization (see below for a more detailed discussion). Black-box optimization algorithms, particularly Bayesian Optimization (BO), have become increasingly prominent over the past decade.\cite{ doi:10.1021/acs.chemrev.2c00798, tom2024self} 
In BO, probabilistic models for predicting reaction yields are built through Bayesian inference with existing data. These models then iteratively guide decision-making throughout the optimization process.
The idea of iterative, closed-loop optimization with ML-based surrogate models is discussed further in Sec. \ref{sec:sdl}.
For condition optimization, these iterative approaches have demonstrated remarkable success in increasingly complex synthetic reaction scenarios.\cite{doi:10.1021/acs.chemrev.2c00798}
At the same time, chemistry-specific challenges, such as the identification of conditions which are ``generally applicable'' to a wide variety of substrates, as opposed to just one or a few model substrates, have inspired algorithmic advances in the field.\cite{doi:10.1126/science.adc8743, Doyle2024} Notably, our work on the Suzuki reaction \cite{doi:10.1126/science.adc8743} led to generally applicable conditions with double the yield of the previous state-of-the-art in the field. 

\subsubsection{Limits and open problems.~~}

While the field of ML-based synthesis planning has seen significant algorithmic advances during the past ten years, its practical utility has remained limited to the development of relatively simple target molecules and short synthetic routes. 
In fact, as of today, expert systems, which involve manually coding reaction types and applicability rules, represent the state of the art in computer-aided synthesis planning. 
In particular, Grzybowski's \textit{Chematica} system (now commercialized as \textit{Synthia})\cite{Chematica} has had impressive experimental applications,\cite{KLUCZNIK2018522} even in complex natural product synthesis,\cite{mikulak2020computational, lin2023computer} or supply-chain-aware synthesis planning.\cite{wolos2022computer, mikulak2024catalyst}
In principle, while ML-based algorithms should be capable of providing similar or superior synthetic routes compared to these expert systems, the current shortcomings can mainly be attributed to deficiencies in the quality and quantity of available synthesis data and algorithmic limitations in extracting structured knowledge from the data. 
We and others have extensively discussed these factors recently.\cite{strieth2024artificial}

Similar data limitations have also been discussed in the context of reaction outcome and reaction condition prediction. Patent data \cite{lowe_2012} and even commercial databases are highly problematic not only because of erroneous, inconsistent or unstructured data reporting: Human biases in the reported experiments, particularly the accumulation of prominent conditions and the lack of low-yielding records, have prevented predictive modelling of reaction yields from literature data.\cite{beker2022machine, anie.202204647} Community-driven, open source data repositories such as the Open Reaction Database\cite{kearnes2021open} represent an essential step towards less biased and more holistic data collection -- but such initiatives require a more digitized mindset in the way data is generated, collected and reported in synthetic organic chemistry laboratories. 

A further consequence of this data deficiency is the lack of representative benchmark problem sets. 
This applies to multi-step synthesis planning, where benchmarks are urgently needed for a more quantitative evaluation of synthesis planning performance. 
Similarly, optimization algorithms for chemical reactivity would benefit from representative benchmarks to evaluate how standard BO algorithms translate to the intricacies of chemical reactivity. 
Most importantly, such benchmarks must reflect real-life problems, as identified by expert chemists, in order to inspire and motivate algorithmic ML advances to tackle the challenges in computer-aided organic synthesis.

\subsection{Structure to physics: simulation and 3D structure}


Machine learning has enabled data-driven solutions to both experimental problems and computational problems.
Whereas organic chemistry emphasizes molecules' 2D molecular graph structure, molecules are also grounded in 3D physical reality by the Schrödinger equation, providing a rich theory of quantum mechanics and statistical mechanics for predicting molecular properties and interactions.
Simulation methods such as density functional theory (DFT) and molecular dynamics (MD) then use this theory to computationally predict molecular properties and interactions.
However, despite continual increases in computing power, these simulations remain computationally costly, which has restricted simulation to small systems at short timescales.
By learning from the results of many simulations, ML offers a unique opportunity to accelerate molecular simulation.

\subsubsection{Neural network potentials.~~}

A fundamental problem in quantum chemistry is: given a molecule represented as a collection of nuclear points in 3D space, solve the Schrödinger equation and predict the total energy and the forces on each atom.
Forces then enable simulation of dynamics forward in time using Newton's equations.
However, solving the Schrödinger equation is complex and computationally costly for molecular systems, and simulating Newton's equations requires forces at every frame of simulation.
For this reason, forces were approximated by simple functions fitted to experimental data, giving rise to the first parameterized force fields such as the Lennard-Jones potential.\cite{jones1924determination}
Semiempirical models incorporated many more experimentally fitted parameters for predicting energy and forces.\cite{christensen2016semiempirical}
These empirical force fields enabled classical molecular dynamics simulations, allowing study of simple proteins.\cite{levitt1975computer}
However, capturing behavior like chemical reactivity requires incorporating quantum effects.
Advances in computer power and faster simulation methods such as density functional theory (DFT) eventually made it possible to solve the Schrödinger equation at every timestep with ab initio molecular dynamics, but at large computational cost.\cite{iftimie2005ab}

A significant shift came with the introduction of neural force fields.
By training neural networks on DFT data to predict energy and forces directly from 3D nuclear coordinates, molecular dynamics could now be propagated at ab initio accuracy at a much lower computational cost.\cite{behler2007generalized}
Since forces must be equivariant to the molecule's rotation -- i.e. if the molecule is rotated, the molecular forces must ``rotate along with it'' -- this motivated the development of equivariant neural architectures to respect this symmetry.\cite{schutt2017schnet,wang2018deepmd,satorras2021n}
Neural force fields have been competitively benchmarked in ML, continually comparing different architectures and methods on several benchmarks. A detailed timeline of development of these equivariant architectures is given in Duval \textit{et al.}\cite{duval2023hitchhiker}
As datasets of energy and forces have grown, such as the Open Catalyst Benchmark,\cite{zitnick2020introduction} neural force fields have started striving for universal applicability.\cite{batatia2023foundation}

\subsubsection{Predicting wavefunctions and electron densities.~~}
An alternative to predicting energies with force fields is to predict the wavefunction or electron density itself. The advantage is that these objects contain energy and the rest of the system's physical observables.
For example, neural networks can be trained to predict the Hamiltonian matrix directly from the nuclear coordinates.\cite{schutt2019unifying,unke2021se}
Diagonalizing the Hamiltonian matrix gives the molecular orbitals, which comprise the wavefunction.
Furthermore, self-consistent field iteration can be initialized using the predicted wavefunction, allowing faster convergence of the quantum chemistry.
Recently, it was shown that neural networks can be trained so that their output satisfies the self-consistency equation, bypassing the need for labels of Hamiltonian matrices.\cite{zhang2024self}

Furthermore, neural networks can be used as ansätze to represent the wavefunction itself directly.
In this case, the network takes as input electron coordinates, and outputs wavefunction amplitude.
Using the same stochastic optimization algorithms, neural wavefunctions can be trained to minimize the variational energy and satisfy the Schrödinger equation.\cite{pfau2020ab, hermann2020deep, glehn2023a, li2023forward, neklyudov2024wasserstein}
This approach has recently been extended to excited states.\cite{pfau2024accurate}

Alternatively, for density functional theory, neural networks can be trained to directly predict charge density given the nuclear coordinates.\cite{fabrizio2019electron, gong2019predicting, fu2024recipe}
ML has also been applied to learn density functionals.\cite{kirkpatrick2021pushing}

\subsubsection{Predicting and generating 3D structure.~~}

Even if fast and accurate force fields were available, many problems rely on finding energetically preferred conformations of molecules.
However, conformational space remains huge and cannot be practically enumerated, especially for large systems like proteins.
Similarly, when modelling chemical reactions, the sizeable conformational search space makes it challenging to identify transition states.
To solve these problems, ML approaches can predict and generate 3D structure directly.

The large conformational search space motivates generative models to navigate this space.
Unconditional generative models such as equivariant diffusion models can generate 3D atomic positions and atom types simultaneously.\cite{hoogeboom2022equivariant}
For the problem of conformer search, which seeks stable 3D configurations for a given molecule, atom types can held constant while generation is conditioned on the 2D molecular graph.
Some approaches generate atom positions freely,\cite{xu2022geodiff} while other approaches generate torsion angles of rotatable bonds.\cite{ganea2021geomol, jing2022torsional}
Recent work has shown that forgoing both torsional and rotational symmetry constraints can yield better results, but at a higher cost.\cite{wang2024swallowing}
A related task known as docking performs conformer search of a ligand inside a protein pocket, as an estimate of binding affinity.
This has also been approached with diffusion models.\cite{corso2022diffdock}

In the problem of crystal structure prediction, the goal is to find the most stable periodic arrangement of atoms for a given composition. While traditional approaches search through all stable configurations of coordinates and lattice vectors to find the lowest energy structure,\cite{pickard2011ab} equivariant diffusion models have found a natural fit for this problem, diffusing both coordinates and lattice parameters simultaneously,\cite{xie2021crystal,jiao2024crystal} while also enforcing space group constraints\cite{jiao2024space} to enhance performance further. Indeed, scaling this diffusion approach to large datasets enabled inverse design to satisfy multiple desired properties simultaneously. \cite{zeni2023mattergen}

In the fields related to the simulation of biomolecules, 3D structure prediction problems are abundant.
The longstanding problem of predicting folded 3D protein structure from protein sequence has, to a certain extent, been solved by AlphaFold2.\cite{jumper2021highly} and related models.
Building on this approach, diffusion models have generated protein backbones represented as sequences of rigid bodies of residues.\cite{yim2023se,bose2023se}
These models have been so successful that they have been used to design proteins satisfying structural constraints, which have been experimentally validated.\cite{watson2023novo,ingraham2023illuminating}
The scope of these diffusion models has expanded to all biomolecules, with methods predicting how proteins, RNA, DNA, and ligands assemble in 3D atomistic detail,\cite{krishna2024generalized,abramson2024accurate} subsuming the task of docking, and hence, promising to become a de-facto conditioning function for drug discovery in the future.

\subsubsection{Enhanced sampling and coarse-grained simulation.~~}

While finding the most stable geometry is useful, truly modelling the thermodynamic interactions between molecules requires sampling the equilibrium distribution of 3D structures.
Equilibrium states follow a Boltzmann distribution with respect to the energy, and generative models which learn this equilibrium distribution are known as Boltzmann generators. \cite{noe2019boltzmann}
Deep generative models are beginning to solve this problem using flow matching,\cite{klein2024equivariant} a variant of diffusion models, and transferability has been demonstrated across many different peptides.\cite{klein2024transferable}
Another approach learns to sample equilibrium distributions by leveraging the Fokker-Planck equation.\cite{zheng2024predicting}

In coarse-graining one typically groups atoms together into so-called beads, which afford lower computational cost and the possibility to capture long timescale events.
However, the forces on these coarse beads then need to be fitted to all-atom forces.
To circumvent this, neural networks can be applied to learn coarse-grained force fields by predicting the gradient of the free energy, rather than the energy, and matching these predicted forces on coarse-grained beads to the all-atom forces.\cite{Wang2019cgnet, husic2020cgschnet, charron2023navigatingproteinlandscapesmachinelearned}
Flow-matching\cite{Koehler2023flowmatchingcg} removes the requirement for all-atom forces, needing only equilibrium samples of coarse-grained beads.
Furthermore, diffusion models can simultaneously learn a generative model and coarse-grained force field.\cite{arts2023two}

While coarse-grained force fields are significantly faster to evaluate than atomistic ones, MD simulations are still limited by having to use femtosecond-level integration time steps.
Alternative methods for equilibrium methods focus on accelerating molecular dynamics to reach long timescales.
This can be done through ``coarse-graining in time,'' which trains generative models to predict the outcome of taking large timesteps.\cite{fu2023simulate, klein2023timewarp}

Lastly, work has been carried out towards extending models to multiple ranges of thermodynamic properties like temperature and pressure.\cite{duschatko2024thermodynamicallyinformedmultimodallearning}
This allows simulation of different environments as well as training on previously unsuitable data.
Adding extra parameters like temperature to the model input, one can add the corresponding derivatives of the coarse-grained free energy function to the loss.
Response properties which are higher order derivatives of the free energy can be computed via multiple backward passes.
Incorporating thermodynamic parameters might be one of the key ingredients to simulate biological or industrial settings in a holistic manner.

For rare-event sampling like chemical reactions and transition state search, methods for sampling transition paths without reaction coordinates have been emerging.\cite{sipka2023differentiable, holdijk2024stochastic}
Alternatively, when datasets of reactants, products, and transition states are available, generative models can be directly trained to generate transition states conditioned on reactants and products.\cite{duan2023accurate,duan2024react}

\subsubsection{Limits and open problems.~~}

While neural force fields can achieve great accuracy, they still require enough training data to cover the entire phase space. Without complete coverage, neural force fields can stumble into unstable dynamics.
One benchmark emphasizes that force fields should be judged by their dynamics, not their force errors.\cite{fu2022forces}

However, these issues may begin to go away as neural forces are trained on ever larger datasets in the quest for universal force fields.
Though ML models are limited by the quality of their data, the fact that new data can be generated by simulation paints a promising picture for data availability and large models.

At the same time, much work remains to reach simulation at large length and time scales. The most significant challenges of proper equilibrium sampling under metastable conditions and the related problem of rare-event sampling also remain areas in need of improvement and, therefore, the focus of many recent efforts.

\subsection{Structure and analysis: spectroscopy and elucidation}
\label{sec:analysis}

One natural yet underexplored area of ML application in chemistry is structure elucidation, which aims to predict 2D or 3D molecular structures from spectroscopic or other analytical data.
Just as computer vision enables computers to perceive the natural world, computational spectroscopy could allow machines to perceive the molecular world through analytical instruments.
The anticipated increase in the synthesis of \textit{de novo} and unknown compounds through advances in experimentation automation drives the need for faster yet accurate structure elucidation to fully support these autonomous molecular and reaction discovery platforms.



\subsubsection{Forward spectral prediction.~~}

The most straightforward approach to data-driven structure elucidation is to store a library of spectra, search for a match in the library for a given spectrum, and then retrieve the corresponding structure.
To increase the coverage of the library, forward spectral prediction can be used to predict spectra given chemical structure.
While physical simulation offers a grounded way to predict spectra, it can be difficult and computationally expensive.
An alternative approach leverages machine learning to predict spectrum from structure, for a variety of types of spectra,
including mass spectrometry (MS),\cite{young2024tandem,young2024fragnnet}
nuclear magnetic resonance (NMR),\cite{paruzzo2018chemical,cordova2022machine}
and ultraviolet-visible spectroscopy (UV-vis).\cite{lupo2023two}
Some frame the forward prediction problem as formula prediction, employing either autoregressive models or a fixed vocabulary of formulas \cite{goldman2023prefixtreedecodingpredictingmass, murphy2023efficientlypredictinghighresolution}; while others focus on subgraph prediction, utilizing recursive fragmentation, autoregressive generation, and deep probabilistic models,\cite{goldman2024generatingmolecularfragmentationgraphs, subset, young2024fragnnet} or incorporate 3D structural information.\cite{3DMolMS, al2024neuralnetworkapproachpredicting} In the context of mass spectra, some methods approximate the spectrum as a sequence of discrete bins with corresponding peak intensities, reducing the problem to a task of regressing the mass spectrum directly from structure.\cite{3DMolMS,young2024tandem} In addition to structure-to-spectrum prediction, another approach involves predicting structure-property relationships by estimating various molecular descriptors -- ranging from scalars (e.g., energy, partial charges) to vectors (e.g., electric dipoles, atomic forces), and higher-order tensors (e.g., Hessian matrix, polarizability, octupole moment) -- and then using these descriptors to predict different spectra, including IR, Raman, UV-Vis, and NMR. \cite{Zou2023}

\subsubsection{Structure elucidation.~~} On the other side is the inverse problem of directly predicting chemical structure from a given spectrum. 
DENDRAL was the first expert system for inferring chemical structure from mass spectra in 1969.\cite{buchanan1969heuristic,lindsay1993dendral}
Chemists also used ML to analyze infrared (IR), nuclear magnetic resonance (NMR), and mass spectra for identifying limited sets of functional groups.\cite{doi:10.1021/ci950102m, CURRY1990213, doi:10.1021/ac60361a029}
While these methods provide helpful structural insights, they are insufficient for fully elucidating molecular structures.

Combining information of many inferred functional groups has enabled structure elucidation.
For NMR data, the molecular structure can be elucidated by first identifying molecular substructures and functional groups,\cite{NMRFG, NMRmixtures,huang2021framework} which are then optimally assembled via beam search over possible configurations or constructed atom-by-atom,\cite{huang2021framework,RL_NMR, D4DD00008K} similar to the approach chemists take when interpreting NMR spectra.
Similar ``reconstruction-by-substructure'' strategies have been employed to varying degrees of structural detail for IR \cite{IR-inverse1,D2SC05892H} and surface-enhanced Raman spectroscopy (SERS). \cite{Tan2024}
However, as the number of atoms increases, this approach quickly encounters combinatorial scaling issues.

Molecular structure elucidation can also be tackled as an end-to-end problem from a deep learning perspective. In this approach, the spectra are tokenized into strings and SMILES strings are predicted; this can be viewed as a machine translation task. This approach has been applied to NMR, IR and tandem MS/MS data,\cite{alberts2023learning, alberts2023leveraging, hu2024accurate,stravs2022msnovelist, Litsa2023} showing more significant promise for scaling to larger chemical systems and {\sl de novo} structure elucidation. The structure prediction problem can also be formulated as an optimization task, e.g. by formulating it as a Markov decision process.\cite{D4DD00008K} If we consider scenarios where we have some prior information about the chemical system at hand, such as chemical formula, known starting materials and reaction conditions, implementing this information as constraints can help the model converge on a solution more efficiently. 

Moving from molecules to crystals, solving the inverse problem for X-ray spectroscopic data such as powder X-ray diffraction (PXRD) and X-ray absorption near-edge structure (XANES) spectra also poses interesting challenges for the machine learning community, where there are unique and underdeveloped opportunities for employing various deep learning models for generalizable crystal system and space group identification.\cite{lai2024end, Salgado2023} Diffusion models have shown particular promise, especially given their successful application to counterpart inverse problems in text-to-image generation. In this context, we can draw parallels between text and spectra and between image generation and crystal structure prediction. \cite{song2022solving, chung2023diffusion} 

In the field of rotational spectroscopy, the challenge of spectral assignment -- i.e. deduce the rotational constants from a densely packed rotational spectrum -- represents one of the earliest application of ML in this domain.\cite{zaleski2018automated} 
This problem is particularly well-suited for deep learning techniques due to the dense yet easy-to-simulate nature of the spectra.
However, the rotational constants alone do not determine the 3D structure of the molecule.
The approach that we recently introduced solves this by inferring 3D structure given incomplete information as molecular formula, rotational constants, and unsigned atomic Cartesian coordinates known as substitution coordinates.\cite{cheng2024determining}

In the realm of structural biology, advances in protein structure prediction have accompanied advances in cryo-electron microscopy. Reconstruction of protein structure from cryo-EM has been tackled using deep generative models.\cite{zhong2021cryodrgn,levy2022amortized}
These methods have progressed to the point of reconstructing biomolecular dynamics from cryo-electron tomography (cryo-ET).\cite{rangan2024cryodrgn} Structure elucidation using CryoEM continues to show day-to-day advances. Advances in data processing have provided incredible gains in resolution\cite{clabbers2024energy} that can only be improved by the use of ML methodologies.

\subsubsection{Limits and open problems.~~}

As with all data-hungry approaches, one key issue remains universal: While simulated spectra can be obtained in large quantities, it is crucial to consider if the model performs well on experimental spectra, which often exhibit more significant variability and inconsistencies. A relevant question to consider is: {\sl Would a more concerted effort by the scientific community to push for the deposition of raw spectral files in open data repositories help advance deep learning applications for automated spectra-to-structure elucidation? }

For inverse spectrum-to-structure elucidation, while autonomous and {\sl de novo} molecular structure determination of pure samples is indubitably essential to facilitate high-throughput reaction optimization and discovery campaigns, it is also crucial to address structure annotation of spectra from complex mixtures, which encompasses both the targeted identification of specific compounds of interest and non-targeted metabolomics. Such mixtures are standard in real-life sample matrices and are essential for various fields ranging from bio-diagnostics to forensics. Success in these tasks is highly contingent on the model’s ability to disentangle and isolate individual molecular spectral signatures from the highly convoluted data. Machine learning excels in handling complex, high-dimensional data, making it well-suited for these challenging tasks.\cite{goldman2023annotating, Baygi2024} In addition, leveraging ML methods to integrate information from multiple spectral inputs could further enhance structure elucidation's accuracy and completeness.

\subsection{Leveraging scale with foundational models for chemistry}


With increasing computational power, machine learning models have been trained on progressively larger datasets.
At scale, ML offers qualitatively different capabilities.
Foundation models are large-scale models that have been trained on a broad spectrum of data and can be applied to a variety of downstream tasks. Several general-purpose foundation models -- such as ChatGPT, Gemini, and Llama -- are typically utilized for language and image generation; many of these are language-only models or models trained on multiple modalities. However, using these models in the chemical domain presents unique challenges, and so many have trained their models from scratch on chemical data, but this is not trivial either. In this section, we will describe the current state of foundation models in chemistry and give our perspective on remaining open questions.

\subsubsection{Transforming knowledge with large language models and agents.~~}

Some of the earliest applications of generative models to chemistry have been via language, which was enabled by the fact that molecules can be represented with strings using SMILES notation.\cite{weininger1988smiles}
Preliminary chemistry language models were trained in an unsupervised manner on SMILES representations,\cite{ross2022large, chithrananda2020chemberta} which learned dependencies between molecular subfragments. More recently, models have also been concurrently trained on other molecular modalities represented by text tokens, such as textual descriptions, scientific papers, synthesis procedures, commonly with autoregressive losses to be able to generate molecule descriptions or structures at inference time.\cite{liu2023multi, pei2023biot5, christofidellis2023unifying, taylor2022galactica, edwards-etal-2022-translation}
\citet{ramos2024review} wrote a comprehensive review detailing 80 chemistry/biochemistry language models to date for further reading. One motivation behind incorporating textual descriptions is that they contain information about functional properties of molecules, which can be useful for improving the embedding representations of molecules that are structurally similar but functionally different, or vice versa. They also enable interaction with models using natural language, which is a more intuitive interface for many users than rigid queries.\cite{kang2024chatmof, yoshikawa2023large}
Additionally, LLMs have been utilized for scientific bibliographic parsing,\cite{Choi2024,gupta2022matscibert,Dagdelen2024} facilitating the extraction of chemical information from existing literature and building knowledge databases. These databases can be used for the fine-tuning of LLMs with the potential to improve the generation and screening capabilities of self-driving labs (Sec. \ref{sec:sdl}).\cite{Buehler2024,MBran2024,kang2024chatmof}
 
However, there still exists a gap in using these models out-of-the-box for discovery tasks or in domain-specific chemistry applications (at least to our knowledge),\citep{taylor2022galactica, ai4science2023impact} one reason being that there is not enough data to train these models in the same way that models like GPT-4 have been trained on web-scale text and images.\citep{achiam2023gpt}
One way to use these chemistry-aware language models is to finetune them on downstream tasks,\cite{jablonka2024leveraging} or plug them into optimization or search frameworks as a way to provide good prior knowledge~\cite{wang2024efficient, kristiadi2024sober, ramos2023bayesian, ma2024llm}. Other works have also begun to explore scaling of both models and data.~\cite{frey2023neural, ross2024gp}

One interesting application of chemistry-aware foundation models has been the development of chemistry agents that can e.g. make use of tools~\citep{schick2023toolformer} necessary for solving chemistry problems, and/or plan chemistry experiments. Some notable examples include ChemCrow, \cite{m2024augmenting} Coscientist, \cite{boiko2023autonomous} our own ORGANA, \cite{darvish2024organa} or ChemReasoner. \cite{sprueill2024chemreasoner} These agents have access to various chemistry-related tools, such as simulators or robots to execute chemistry experiments, and use an LLM (such as GPT-4) as a central orchestrator to decide when and how to use these tools to accomplish a user-specified goal. One longer-term goal of such agents is to develop scientific assistants that can help beyond calculating and executing to do more complex reasoning and planning by generating and refining hypotheses on their own. This has been extended to other research domains by the AI Scientist, which demonstrates autonomous machine learning research by executing experiments and writing a research paper.\cite{lu2024aiscientistfullyautomated}

These research areas are in their infancy, so several open questions remain, including: (1) How do we effectively evaluate chemistry-aware LLMs/agents? (2) What are the use cases for these models in practice for chemists? 
Effective model evaluation mainly depends on developing meaningful tasks, which is currently an open problem both in terms of dataset scale and breadth. There already do exist several benchmarks in this space,\citep{wu2018moleculenet, huang2022artificial} which is a good start but there is room to improve them in terms of data quality and task objectives.\citep{practicalcheminformaticsNeedBetter} More recent benchmarks have been released that are closer to real-world applications,\cite{nigam2023tartarus,mirza2024large, laurent2024lab} and also platforms such as Polaris have made it easier for researchers to have faster access to a wide array of datasets.\cite{polarishubPolaris}
The issue with using sub-optimal benchmarks in this field has been exacerbated by the current climate in machine learning in that benchmarks are mainly used to show that a new method achieved better performance than the current state-of-the-art, without human understanding of why it improved. This is also an excellent opportunity for collaboration between chemists and the ML domain expert communities.

Language-based foundation models have also been used in other applications, including knowledge graph generation\cite{venugopal2024matkg} and  knowledge extraction from chemical literature,\cite{bran2024knowledge, ai2024extracting, zheng2023chatgpt, schilling2024text} including our own work on reaction diagram parsing,\cite{MERMES} which is a difficult task. These efforts are essential for creating structured databases of experimental procedures, which can contribute to existing repositories such as the previously-mentioned Open Reaction Database.\cite{kearnes2021open}


\subsubsection{Foundational physical models.~~} 

While language-only foundation models are receiving a lot of attention in chemistry, it has been shown that language might not be the sufficient modality, especially in settings where 3D geometry matters. For example, \citet{alampara2024mattext} showed that language models are not enough to encode structural information needed to represent specific material properties.

However, language models are not the only foundation models developed in the biochemical sciences. Several models have been built to universally approximate force fields and predict structures for any molecule, material, or protein.\cite{duignan2024potential, batatia2023foundation, liao2024equiformerv, merchant2023scaling, yang2024mattersim} Perhaps the most famous example is AlphaFold2 for protein structure prediction\cite{jumper2021highly} and, more recently, AlphaFold3,\cite{abramson2024accurate} which given any set of 2D biomolecules, predicts how they might assemble in 3D. To our knowledge, these models still outperform any sequence-based protein prediction models for many structural and functional tasks, especially in cases where input sequences do not have homologues in the training data.\citep{van2024fast}

Another impressive example is the recent foundation model MACE-MP-0, built with the MACE equivariant architecture.\cite{batatia2022mace, batatia2023foundation} MACE-MP-0 was trained on 150 thousand inorganic crystals. After a small number of task-specific examples for fine-tuning, it can be used as a force field in simulations on a wide variety of tasks, even seemingly unrelated ones such as small protein simulations. Notably, intermolecular interactions seem somewhat fuzzy in the MACE-MP-0. For example, in the aforementioned protein simulation, the model was able to capture hydrogen transfer, which is a remarkable achievement. However, the authors also opted to include D3 dispersion borrowed from classical computational chemistry, pointing to the fact that the model still needs some help to predict long-range interactions.
Foundational force fields have continued to scale, with industry research labs training neural force fields on ever-larger data, such as GNoME\cite{merchant2023scaling} and MatterSim.\cite{yang2024mattersim}


One key takeaway from these types of models is that structural information should not be ignored depending on what downstream tasks the model will be applied to, and that training  models on broad, large-scale datasets (i.e., going beyond training a simple model on a single prediction task, which was the norm even a couple years ago) can help generalize better to more downstream settings. We suspect that scaling along multiple modalities concurrently is critical for building the best foundation model in chemistry -- namely, training models on as many modalities as possible, such as 3D structure information, text, and spectral information. \cite{zhou2023unimol}







\subsubsection{Limits and open problems.~~}
In the case of the domain sciences, we are not as privileged as in the domain of natural language or images, which already has internet-scale data available. Scientific data is scarce; every data point must be an experiment or a high-quality simulation. If simulations are employed, the model must find a way to translate their results to specific experimental conditions. We suspect that {\sl universal} models across chemistry are still a decade away and will perhaps be a moving target as humans continue to demand more of them. This is analogous to the problem of widening highways \cite{weingart2023widening} where many analysts have shown that as soon as a road is widened, the additional created demand due to its availability makes the highway full of traffic immediately. 

\subsection{Closed-loop optimization and self-driving labs}
\label{sec:sdl}


%

\subsubsection{Self-driving laboratories.~~}

As ML applications continue to evolve, the necessity and scarcity of high-quality data become increasingly apparent. The advent of chemical digitization\cite{Raghunathan2021,Wigh2022} and advances in ML\cite{aldossary2024silico,Meuwly2021} have laid the groundwork for combining ML with automated data generation through robotic experimentation. This synergy has given rise to the concept of the self-driving laboratory (SDL).\cite{tom2024self} SDLs are primarily composed of two critical components: automated laboratory equipment and experimental planners, both of which leverage ML techniques to enhance their functionality.\cite{tom2024self} The ultimate goal is to autonomously execute the scientific method, encompassing hypothesis generation (ML), hypothesis testing (experimentation), and hypothesis refinement (ML), potentially allowing for the exploration of vast design spaces in a data-efficient manner.

Significant advancements in automated laboratory equipment have been achieved by integrating ML with computer vision,\cite{Wang_2023} leading to the concept of ``general chemistry robots.''\cite{Burger2020} These ML-trained robots can make decisions based on external feedback, enabling the dynamic automation of chemical operations traditionally performed by human chemists.\cite{Nakajima_2022,Kennedy_2019,Huang2021} Given the inherent challenges in training robotic equipment for active decision-making based on external feedback, a notable innovation in this area is the use of digital twins---virtual replicas of laboratory setups—that provide a robust framework for accelerating the training of robotic ML models.\cite{Klami_2022} These digital twins simulate chemical scenarios with high fidelity,\cite{beeler_2023} creating a realistic feedback loop that accelerates the model's learning process.

On the experimental planning side, heuristic techniques\cite{Bezerra2016,snobfit,Lucia1990} are being progressively replaced by ML optimization algorithms. When combined with chemical digitization,\cite{Walters2020} these optimization techniques can identify target chemicals and optimize reaction conditions while significantly reducing the number of experimental steps required.\cite{Taylor2023} Among the various ML optimization techniques,\cite{Zhou2017,Jastrebski} Bayesian optimization\cite{Hse2018,Hse2021,Hickman2023} has gained particular prominence in experimental chemistry due to its success in chemical applications.\cite{shields2021bayesian} Machine-learning-based surrogate models, which predict the properties of chemicals and reactions,\cite{Eyke2021,Oliveira2022,Dara2021} have been instrumental in this success, with documented examples in both process optimization and materials discovery.\cite{strieth2024delocalized}

Moreover, the rise of LLMs has further enhanced the auxiliary components of SDLs. LLMs have been effectively used to create human-machine interfaces that bypass traditional coding,\cite{darvish2024organa} enabling more natural communication between chemists and laboratory systems---a significant advantage for users who may not be well-versed in coding or data processing.\cite{darvish2024organa,clair,Yoshikawa2023}


\subsubsection{Limits and open problems.~~}
As discussed by us recently,\cite{Seifrid2022} the challenges facing SDLs can be broadly categorized into two areas: motor (hardware-related) and cognitive (AI-related).

{\bf Motor challenges.} The primary hardware challenges stem from the human-centric design of chemical instruments and the lack of seamless interconnection between existing automated modules. As a result, most SDLs operate semi-automatically, requiring human intervention for tasks such as sample transfer, maintenance, and troubleshooting. Various solutions have been proposed to address these issues, including deploying mobile robots for sample transfer\cite{Burger2020} and adapting general-purpose robots to perform chemical tasks or operate instruments originally designed for human use.\cite{Knobbe_2022,Yoshikawa2024,Jiang2023} However, many of these methods rely on traditional algorithms that require static calibration, which is not well-suited to the dynamic nature of SDLs. While computer vision coupled with AI has been proposed as a solution, laboratory equipment, particularly glassware, continues to present significant challenges that are continuously being addressed.\cite{Xu2021}

{\bf Cognitive challenges.} Cognitive challenges primarily arise from the difficulty in developing models that can accurately estimate the chemical output of the system. This limitation restricts the use of more general generative models, effectively reducing the amount of chemical space that experimental planners can explore. When combined with the aforementioned motor challenges, another issue becomes apparent: SDLs often operate in low-data regimes. Predictive and generative machine learning models typically require large datasets to make meaningful predictions. While generative models can be trained on existing data,\cite{Anstine2023,wang2024efficient} deploying predictive algorithms in such low-data regimes remains a significant challenge.

{\bf Auxiliary component challenges.} Regarding the auxiliary components of SDLs, the incorporation of LLMs shows promise in automating workflow creation\cite{clair} and improving human-machine interfaces. However, further research is needed to ensure the safety and reliability of these processes. Additionally, while integrating bibliographic extraction into SDLs can enhance model development, its effective integration with predictive models remains an unresolved issue.

A final challenge to be addressed in the field of SDLs is the economy of scale of their development. The more SDLs the community builds, the easier it will be to build the next ones. Hence, the democratization of low-cost SDLs is crucial for the advancement of the field.\cite{lo2024review}

\section{Problems meet methods: a machine learning perspective on solving chemical problems}


There is already a wealth of resources on how to apply the specifics of machine learning in several books, reviews, and internet resources.\cite{janet2020machine, keith2021combining, artrith2021best, white2022deep} In this section, we provide a  high-level perspective of how ML researchers and communities view and tackle problems.
To start, we reclassify the diverse chemical problems introduced above as instances of well-established ML problems.
To elaborate the ML perspective, we gather common themes and practices in the ML community and examine them in light of application to chemistry, highlighting points to consider related to benchmarking, the role of domain knowledge, and community values.

\subsection{The toolbox of machine learning}

ML provides a toolbox of algorithms and theory for solving problems using data.
ML has formalized a set of well-defined problems to solve diverse tasks in language, vision, audio, video, tabular data, scientific data, and other domains.
Each problem establishes a set of input requirements and a desired goal, which has proved helpful for empirically benchmarking and theoretically analyzing different algorithms under a common framework.
In Table \ref{tab:toolbox}, we lay out significant ML problems with their expected inputs and goals and reclassify different chemical problems as instances of these ML problems.

\textbf{Regression and classification} aim to predict labels $y$ from inputs $x$, given a dataset of paired data. Labels can be one-dimensional, such as in predicting properties, energy, or yield, but also high-dimensional, such as the ML regression problems related to force fields, spectra prediction, and segmentation.
When data is small and tabular, gradient boosting machines such as XGBoost\cite{chen2016xgboost} often perform well.
Gaussian processes also work with small data and provide good uncertainties for use in Bayesian optimization.\cite{tom2023calibration}
However, deep neural networks are the algorithm of choice for high-dimensional, complex data like images, text, and molecules.
The choice of neural network architecture is informed by the problem's constraints: graph neural networks for 2D graphs and equivariant architectures for 3D data.
Relatively recently, transformers\cite{vaswani2017attention,lin2022survey} have revolutionized modelling of language,\cite{vaswani2017attention} images,\cite{dosovitskiy2020image} graphs,\cite{ying2021transformers} and 3D molecules.\cite{jumper2021highly,liao2024equiformerv}

\newcommand{\tabitem}{~~\llap{\textbullet}~~}

\begin{table}[t]
\small
\centering
  \caption{~A toolbox of machine learning}
  \label{tab:toolbox}

\makebox[\textwidth][c]{
\begin{tabular}{l l l l l}
\hline
ML problem & Input & Goal & Chemical problems & Algorithms \\
\hline
\begin{tabular}[t]{@{}l@{}}
Regression and\\
classification
\end{tabular} &
\begin{tabular}[t]{@{}l@{}}
Paired data\\
$\{(x,y)\}$
\end{tabular} &
predict $\hat{y} = f(x)$ &
\begin{tabular}[t]{@{}l@{}}
\tabitem Property prediction\\
\tabitem Neural network potentials\\
\tabitem Yield prediction\\
\tabitem Proxies for fast prediction\\
\tabitem Spectra prediction\\
\tabitem Figure segmentation\\
\tabitem (3D structure prediction)\\
\end{tabular} &
\begin{tabular}[t]{@{}l@{}}
\tabitem Classical machine learning: \\
\quad linear regression, random forests, \\
\quad support vector machines,\\
\quad gradient boosting machines\\
\tabitem Gaussian processes\\
\tabitem Neural networks\\
\tabitem Graph neural networks\\
\tabitem Equivariant neural networks\\
\tabitem Transformers
\end{tabular}  \\

\hline

\begin{tabular}[t]{@{}l@{}}
Generative\\
modelling
\end{tabular} &
\begin{tabular}[t]{@{}l@{}}
Dataset $\{x\}$,\\
optional \\
conditioning $\{y\}$
\end{tabular} & 
\begin{tabular}[t]{@{}l@{}}
draw samples\\
$x \sim p(x)$ or\\
$x \sim p(x|y)$
\end{tabular} &
\begin{tabular}[t]{@{}l@{}}
\tabitem Conformer search\\ 
\tabitem Docking\\ 
\tabitem Crystal structure prediction\\ 
\tabitem Transition state search\\ 
\tabitem Structure elucidation\\ 
\tabitem Forward synthesis prediction\\ 
\tabitem (Molecular design)
\end{tabular} & 
\begin{tabular}[t]{@{}l@{}}
\tabitem Variational autoencoders\\ 
\tabitem Generative adversarial networks\\ 
\tabitem Normalizing flows\\ 
\tabitem Autoregressive models\\ 
\tabitem Denoising diffusion \\
\quad and flow matching
\end{tabular} \\

\hline

Sampling & 
\begin{tabular}[t]{@{}l@{}}
energy $E(x)$
\end{tabular} &
\begin{tabular}[t]{@{}l@{}}
draw samples\\
$x\sim p(x)\propto e^{-E(x)}$
\end{tabular} & 
\begin{tabular}[t]{@{}l@{}}
\tabitem Equilibrium sampling\\ 
\tabitem Transition path sampling\\
\tabitem Molecular design\\ 
\end{tabular} & 
\begin{tabular}[t]{@{}l@{}}
\tabitem Markov chain Monte Carlo\\
\tabitem Sequential Monte Carlo\\
\tabitem GFlowNets
\end{tabular} \\

\hline

\begin{tabular}[t]{@{}l@{}}
Gradient-based\\ 
optimization
\end{tabular} & 
loss $\mathcal{L}(\theta)$ & 
\begin{tabular}[t]{@{}l@{}}
optimal\\
parameters $\theta^*$ 
\end{tabular} & 
\begin{tabular}[t]{@{}l@{}}
\tabitem Neural wavefunctions\\ 
\tabitem Physics-informed \\
\quad neural networks \\
\tabitem Differentiable simulation\\
\tabitem (Molecular design)\\
\end{tabular} & 
\begin{tabular}[t]{@{}l@{}}
\tabitem First-order: (stochastic) \\
\quad gradient descent, Adam\\ 
\tabitem Second-order: K-FAC\\
\end{tabular} \\

\hline

\begin{tabular}[t]{@{}l@{}}
Black-box\\ 
optimization
\end{tabular} & 
Oracle $f(x)$ & 
optimal $x^*$ &
\begin{tabular}[t]{@{}l@{}}
\tabitem Reaction and \\
\quad process optimization\\ 
\tabitem (Molecular design)
\end{tabular} &
\begin{tabular}[t]{@{}l@{}}
\tabitem Bayesian optimization\\ 
\tabitem Bandit optimization\\ 
\tabitem Reinforcement learning\\ 
\tabitem Genetic algorithms
\end{tabular} \\

\hline

\begin{tabular}[t]{@{}l@{}}
Agents\\
\end{tabular} & 
\begin{tabular}[t]{@{}l@{}}
Environment \\
of states $\{s\}$, \\
actions $\{a\}$, \\
transitions, \\
and reward $R(s)$
\end{tabular} & 
\begin{tabular}[t]{@{}l@{}}
draw actions \\
from optimal \\
policy $a \sim \pi^*(s)$
\end{tabular} &
\begin{tabular}[t]{@{}l@{}}
\tabitem Extracting literature data\\ 
\tabitem Executing simulations \\ 
\tabitem Question answering \\ 
\tabitem Synthesis planning
\end{tabular} &
\begin{tabular}[t]{@{}l@{}}
\tabitem LLM prompting frameworks \\
\tabitem Reinforcement learning \\
\end{tabular} \\

\hline
\end{tabular}
}
\end{table}

\textbf{Generative modelling} aims to draw samples $x$ from a distribution $p(x)$ defined by a dataset $\{x\}$. Unconditional generative modelling tries to match the data distribution. Conditional generative modelling takes a label or prompt $y$ and tries to learn the conditional distribution $p(x|y)$, blurring the line between unsupervised and supervised learning.
While unconditional generative modelling is rarely valuable for chemistry, conditional generative modelling is ideally suited to inverse problems or one-to-many problems. This is the case for conformer search (one 2D structure for many 3D conformers), structure elucidation (one signal could be consistent with multiple molecules), or forward synthesis prediction (given reactants, many products might be possible). Generative models are a natural fit for their ability to produce multiple quality answers to a question. On the other hand, regression will average over all the possible answers, which may not be a quality answer itself.
Whereas AlphaFold2 \cite{jumper2021highly} used regression to predict one 3D structure given one sequence, AlphaFold3 \cite{abramson2024accurate} used diffusion models to predict multiple biomolecular assemblies for the same input structures.
While many generative model classes exist, such as variational autoencoders,\cite{kingma2013auto} generative adversarial networks,\cite{goodfellow2014generative} and normalizing flows,\cite{rezende2015variational} the dominant ones today are autoregressive models for language\cite{brown2020language} and diffusion/flow matching models for perceptual data like images.\cite{ramesh2022hierarchical}
In chemistry, this translates to chemical language models of SMILES\cite{ross2024gp} and diffusion models of 3D molecular structure.\cite{abramson2024accurate}
Both approaches rely on gradual generation via iterative prediction by a neural network, usually a transformer.
Because an unconditional generative model learns to reproduce a data distribution, which may be a large amount of plentiful unlabeled data, training a generative model can also be thought of as compressing all this data into the network's weights, imbuing a notion of understanding.
Tasks such as sampling and agent behaviour can then build on this understanding.

\textbf{Sampling} also aims to draw samples from a distribution but is distinguished from generative modelling because it only permits access to an energy function $E(x)$, which defines an unnormalized probability density $p(x)\propto e^{-E(x)}$.
No dataset is provided, so one cannot simply train a generative model.
Furthermore, generating a dataset in the first place would require drawing samples.
In addition, the energy function is often computationally costly to evaluate.
For these reasons, sampling problems are among the most difficult in ML and computational chemistry.
Numerous sampling algorithms exist in the literature, with many originating from statistical mechanics, such as Markov chain Monte Carlo (MCMC)\cite{metropolis1953equation} and Langevin dynamics.\cite{parisi1981correlation}
These traditional methods are beginning to incorporate ideas from modern machine learning, such as drawing inspiration from diffusion models for MCMC, \cite{chen2024diffusive} or incorporating learnable components into sequential Monte Carlo.\cite{zhao2024probabilistic}
Some methods learn a bias potential to do transition path sampling,\cite{holdijk2024stochastic} while other methods turn diffusion models into samplers which can solve combinatorial optimization problems.\cite{sanokowski2024diffusion}
Sampling methods are key to solving equilibrium sampling problems, which are necessary for predicting the thermodynamics and kinetics of many chemical processes.
Generative models can be used as components of sampling algorithms,\cite{rotskoff2024sampling} such as in Boltzmann generators,\cite{noe2019boltzmann,zheng2024predicting} which train both by energy and by example.
Boltzmann generators have also begun to leverage generative models, transferring learning between different examples.\cite{klein2024transferable}
Generative Flow Networks\cite{bengio2021flow} (GFlowNets) solve this sampling problem by learning to distribute flow in a generative graph, with a unique strength for generating diverse, discrete data.
Indeed, a growing body of literature has applied GFlowNets to molecular and materials design problems.\cite{jain2022biological,hernandez2023crystal,koziarski2024rgfn,zhu2024sample}

\textbf{Gradient-based optimization} seeks to optimize a smooth loss function $\mathcal{L}$ with respect to parameters $\theta$, which is used to train the neural networks used to solve nearly all of the other ML problems. To do so, machine learning has developed a suite of optimization algorithms such as (stochastic) gradient descent, Adam,\cite{kingma2014adam} and second-order methods such as K-FAC\cite{martens2015optimizing} which use second-derivative information.
Machine learning frameworks such as PyTorch,\cite{paszke2019pytorch} JAX,\cite{jax2018github} and Tensorflow\cite{tensorflow2015-whitepaper} have implemented automatic differentiation with GPU acceleration, making it easier to optimize neural networks.
The fact that neural networks can be optimized so well has motivated the use of neural networks as ansätze for finding wavefunctions to satisfy the Schrödinger equation.\cite{pfau2020ab} This approach, in turn, is an instance of a physics-informed neural network (PINN),\cite{raissi2019physics} which seeks neural network solutions to PDEs by using the PDE itself as a loss function.
Automatic differentiation also enables propagating derivatives through simulation, which can learn potentials for pairwise interaction,\cite{wang2023learning} bias potentials for transition path sampling,\cite{sipka2023differentiable} and perform inverse design.\cite{vargas2023inverse}

\textbf{Black-box optimization} methods try to optimize an oracle function $f(x)$ in a derivative-free manner with as few oracle calls as possible.
This is the case in many experimental problems such as optimizing reaction parameters for yield,\cite{shields2021bayesian} device processing parameters for performance,\cite{osterrieder2023autonomous} or liquid handling parameters.\cite{velasco2024optimization}
To solve these problems with high sample efficiency, algorithms like Bayesian optimization and bandit optimization are applied.
When sample efficiency is not a concern, families of algorithms such as reinforcement learning and metaheuristic optimization like genetic algorithms can also be applied.\cite{tripp2023genetic}
Black-box optimization can also be treated as an instance of sampling, where the target distribution is concentrated around the global optimum.

\textbf{Agents} solve complex multistep problems within an environment.
An environment defines possible states $s$, actions $a$, transitions between states, and a reward function $R(s)$.
For example, retrosynthesis planning\cite{segler2018planning} has molecules as states, chemical reactions as actions, and yield and cost as reward functions.
Planning problems such as retrosynthesis planning or robotic motion planning\cite{skreta2024replan} are naturally solved by agent behaviour, and standard algorithms to learn optimal agent behaviour are known as reinforcement learning.
Because reinforcement learning has poor sample efficiency, a common approach is to initialize agents from generative models:
Helpful assistants such as ChatGPT were initialized as large language models pretrained on internet-scale text, followed by finetuning to maximize a reward of satisfying human preferences.\cite{ouyang2022training}
Prompting frameworks are a rapidly emerging set of methods for augmenting these agents' capabilities, allowing them to reason step-by-step,\cite{wei2022chain} use tools,\cite{schick2023toolformer} retrieve information,\cite{gao2023retrieval} and execute code,\cite{gao2022pal} and to continually repeat these steps.\cite{yao2022react}

\subsubsection{The benefits of a toolbox.~~}

A shared problem interface enables clear and broad benchmarking of many different algorithms.
One example can be seen in Table 1 of Song \textit{et al.},\cite{song2023consistency} who propose a new class of generative models and extensively compares their method to 27 different generative models of different classes on the same dataset and benchmark.

Each of these ML problems also has its own theoretical foundations. Mathematical theory can analyze algorithms for proofs of convergence or properties when converged, providing explanations of why certain methods work better than others. The shared problem interface also allows analysis to determine when one method is the same as another or which methods are more general than others, which helps unify a diverse literature.

\subsubsection{Tools can be stacked on top of each other.~~}

ML problems are also intertwined with each other.
Generative models, like diffusion models, use neural networks trained to regress denoising steps. Agents are built on top of generative text models, while the core of the generative model itself is a neural network predicting the next token. All these networks are trained using stochastic optimization methods like Adam, while black-box optimization is used to choose network hyperparameters.
Sampling algorithms, black-box optimization, and agents can also incorporate generative models trained on previous data, improving the data generation quality.


The problems enumerated in Table \ref{tab:toolbox} are not an exhaustive list. Other problems include uncertainty quantification, which is helpful in Bayesian optimization\cite{griffiths2024gauche} and active learning,\cite{ang2021active} 
federated learning for combining industrial pharmaceutical data while preserving privacy,\cite{heyndrickx2023melloddy}
representation learning for generally applicable molecular descriptors,\cite{zaidi2022pre}
causal learning, retrieval, and compression.

\subsubsection{Picking the right tool for the job.~~}

While the tools of ML are powerful, they provide the most mileage when used for the right job.
For example, as mentioned previously, generative modelling is more naturally suited for one-to-many problems such as 3D structure prediction.
Gradient-based optimization is applicable when the loss function is differentiable and fast to evaluate, such as for optimizing neural networks, but not necessarily for optimizing molecular structure.
While molecular design is often viewed as a black-box optimization problem, it can be argued that sampling is the proper framework for molecular design: Discovery as a multiobjective problem seeks many diverse but quality hits, whereas black-box optimization tends to locally focus on the best solution seen so far.\cite{jain2023gflownets}
Molecular design cannot be solved by generative modelling alone because generative models learn the distribution of a given dataset. In contrast, molecular design seeks exceptional candidates outside the known data distribution.

In chemistry, there is a tendency to treat problems as a search, like finding a needle in a haystack. 
Traditional docking approaches search for all feasible ligand positions, while crystal structure prediction exhaustively searches for all atom arrangements.
Molecular design by virtual screening assumes there will be sufficiently good needles in a haystack of large virtual libraries.
A search-based perspective is useful when available resources are sufficient to exhaustively model a space, which may be necessary to show that no good solutions exist.
However, for many applications, an exhaustive search is overkill.
Imagine trying to write an essay by searching over the space of all possible English texts.
A helpful exercise is to ask whether a search problem has the data and algorithms available to be reframed as a generative modelling or sampling problem.

\subsection{Themes and practices in the ML community}

Solving chemical problems can be aided by both high-level perspectives and community practices.
To contextualize ML perspectives on algorithm development, we describe common themes and practices in the ML community, such as benchmarking, extreme interdisciplinarity, and the bitter lesson of deep learning. All of these are expanded below.

\subsubsection{The role of benchmarking.~~}

Benchmarking plays a crucial role in the ML development process, driving the continuous improvement of models and methods.
The ML community highly values methods that improve on the state of the art. With at least three major computer science conferences annually (NeurIPS, ICML, and ICLR), incremental advances are frequent. These minor, iterative improvements on established benchmarks often accumulate to gain significant performance gains over time. For researchers, benchmarks provide a clear metric for assessing which components of a model most affect performance, enabling more focused and impactful developments.


A prominent feature of ML research is the use of leaderboards, where proposed methods are ranked based on their performance against established benchmarks. Papers must either advance or be competitive with the state of the art to be accepted at major conferences. This process has driven notable progress in various domains, from image classification \cite{deng2009imagenet} and machine translation\cite{bojar2014findings} to image generation,\cite{heusel2017gans} and even solving Olympiad math problems.\cite{hendrycks2021measuring} Leveraging this mechanism, the Open Catalyst Project\cite{zitnick2020introduction, chanussot2021open, tran2023open} set a benchmark for neural network potentials to relax organic adsorbates on metal surfaces. This project provided a dataset much larger than encountered before, which motivated the continual development of more powerful equivariant architectures. From 2020 to 2023, the success rate of predicting adsorption energy grew from 1\% to 14\%, with current models now becoming useful in predicting adsorption. \cite{lan2023adsorbml,metademolabOpenCatalyst}
Another benchmark called Matbench Discovery\cite{riebesell2024matbenchdiscoveryframework} has initiated an arms race of neural force fields on the industry level.

However, while benchmarking is a powerful tool, it is essential to be critical of its applicability to chemistry. Domain experts are uniquely positioned to define practical benchmarks that can translate to real-world outcomes in the lab.\cite{gao2022sample, nigam2023tartarus} Too often, ML literature presents problem settings that, while optimized for computational performance, may be unrealistic for experimental validation. This misalignment can lead to a scenario where the focus shifts from solving the actual problem to merely advancing ML techniques. As methods mature and benchmarks become saturated, new, more relevant benchmarks must arise.


Ultimately, defining and framing problems for ML researchers is a critical task. It involves proposing important questions and calls to action in a way that is accessible to the broader ML community. By doing so, chemists can guide the development of ML tools more likely to have practical applications in experimental research.
While creating datasets and benchmarks can be seen as rote work, it can spur progress on difficult problems by leveraging community efforts of the ML community.
Suppose a chemical problem can be crystallized and packaged into a clearly and appropriately benchmarked ML problem.
Chemists can now wonder: What new problems now become possible to solve, if these old tasks can be solved with significantly greater speed or accuracy?
There are many more scientific questions in the vast set of exciting areas to work in chemistry and materials.




\subsubsection{Interdisciplinary: the effect of chemistry on ML.~~}
\label{sec:interdisciplinary}

Whereas benchmarking iterative improvements is a mainstay of methods-driven ML in the computer science community, an alternative approach to innovation leverages the extreme interdisciplinarity of the ML community.
ML has been applied in fields as diverse as health, agriculture, climate, conservation, physics, and astronomy.
We recently suggested application-driven ML\cite{rolnick2024application} as an emerging paradigm that evaluates success based on real-world tasks in diverse areas, with methods and evaluations informed and contextualized by domain knowledge.
Application-driven innovation acknowledges the impact of incorporating tasks from these diverse areas on the development of machine learning.
New tasks motivate new algorithms.

For chemistry, the development of graph neural networks was driven by the need to model molecular graphs.\cite{duvenaud2015convolutional, gilmer2017neural}
This led to practical advances in modelling other graph data like social networks, citation networks, computer programs, and databases. Graph machine learning in turn made theoretical advances, particularly in analyzing the expressivity of GNNs through the Weisfeiler-Lehman test.\cite{xu2018powerful,delle2024three}
In addition, the need for neural networks to respect rotational symmetries of 3D space motivated the development of equivariant architectures.\cite{thomas2018tensor}
All these methodological developments in respecting symmetries have been unified with a theory of geometric deep learning,\cite{bronstein2021geometric} which shows how convolutional neural networks, graph neural networks, and transformers are actually tightly related.

Beyond theory and methods, ML researchers are also excited for the potential of ML to help tackle real-world problems like global health and climate change.
This has manifested as a great eagerness to learn, as evidenced by the proliferation of blog posts,\cite{ai4science101MolecularSimulation} teaching material,\cite{white2022deep} and online reading group communities with recorded talks.\cite{valencelabsPortal}
Several workshops which focus on ML applications to chemistry are offered at main ML conferences such as NeurIPS,\cite{googleAI4MatNeurIPS2024,mlsbMachineLearning,genbioworkshopGenBioNeurIPS} ICML,\cite{icmlcompbioCompBioWorkshop,ai4sciencecommunityScienceScaling} and ICLR.\cite{googleMLDD2023,ml4materialsICLR2023}
This wide availability of resources also reflects the value of openness in the ML community.
Conference papers are published freely, preprints are emphasized, and sharing code is expected.
Conferences even have a track for accepting blog posts.\cite{iclrblogpostsAboutICLR}

When speaking to ML researchers, be patient with their initial assumptions. Often, several assumptions are made in the ML literature, which ultimately pan out to lose applicability when applied to actual experiments. This occurs in molecular design neglecting the synthesizability of molecules,\cite{gao2020synthesizability} or in reaction prediction neglecting the reaction conditions.\cite{schneider2016big} This reflects the different values and assumptions reviewers make in a distinct field. It is easy to view this and dismiss those approaches as naïve, and it is good to make these criticisms. But let us not throw the baby out with the bathwater: We should ask, if these additional assumptions were taken care of, could this approach help solve our problem? As ML practitioners come from different backgrounds, they will not immediately understand jargon assumptions and experimental setups in chemistry. But they are eager to learn.









\subsubsection{The bitter lesson: Balancing scalability with domain knowledge.~~}
\label{sec:bitter}

The advent of AlexNet\cite{NIPS2012_c399862d} marked the beginning of the deep learning revolution, showcasing how neural networks, when trained using the computational power of GPUs, could classify images with much better accuracy than models based on hand-designed features. The power of computational scale was made explicit with the observation of neural scaling laws,\cite{kaplan2020scaling} which empirically but reliably predict how model performance improves as compute, data, and parameter counts increase. These scaling laws motivated the GPT series of language models,\cite{radford2019language,brown2020language,achiam2023gpt} which ultimately led to advanced applications like ChatGPT.

In light of scaling laws, we should be careful when imposing our domain knowledge when designing algorithms. The ``bitter lesson'' in ML cautions against relying too heavily on domain knowledge when designing algorithms.\cite{sutton2019bitter} While hand-crafted, domain-specific design choices can offer short-term improvements, approaches that better leverage computational scale often outperform them in the long run.
Across domains like text, images, speech, chess, and Go, approaches which rely on human intuition and inductive bias have been replaced by ``brute-force'' approaches that can take advantage of exponential increases in computing power provided by Moore's law.

As chemists, it is joyful to develop methods that are informed by our chemical knowledge, such as by injecting quantum chemistry descriptors into regression,\cite{li2024quantum} or by imposing physical constraints on the system.
However, we should remind ourselves that our human understanding of a problem does not directly translate into being able to design algorithms that solve this problem.
Despite extensive knowledge of linguistics in ML research, models like ChatGPT were not realized until researchers trained on massive datasets.

The power of scale can be fearful. Even beloved assumptions like enforcing equivariance in neural networks have been challenged by recent work: Methods like probabilistic symmetrization\cite{kim2023learning} and stochastic frame averaging\cite{duval2023faenet} have shown that imposing architectural constraints is not strictly necessary, while models like AlphaFold3\cite{abramson2024accurate} and Molecular Conformer Fields\cite{wang2024swallowing} have demonstrated that shown that models trained with randomly rotated training examples can automatically learn rotation equivariance, but at the cost of higher computation and longer training time.

At the same time, the present-day has limited scale and data.
For example, expert systems with reaction rules are still the most effective approach for synthesis planning today,\cite{Chematica} perhaps owing to the difficulty of gathering reaction data.
In addition, one can discard even more inductive bias and train language models to generate 3D molecular structure directly as .xyz files, as we did recently,\cite{flam2023language} and it can compare favourably with more hand-tailored methods for crystal structure prediction.\cite{gruver2024finetuned}
Yet, as \citet{alampara2024mattext} showed, current language models cannot encode geometric information needed to represent specific material properties.

Therefore, the bitter lesson does not mean that imposing inductive bias on algorithms is never good.
An optimal balance must be chosen between leveraging computational power and domain expertise.
This is especially critical in chemistry: Unlike domains like language and images, which are available at internet-scale, chemical data is scarce and costs real-world experiments to obtain.
It is crucial to design algorithms which use this limited data most efficiently.
Hand-designed algorithms can enable better predictions and faster simulations in the near-term, which can bootstrap data generation towards ultimately reaching the scale of data required for foundation models.

Another critical role of domain knowledge is determining the appropriate concept of a problem. Should we model it from first principles, like physics-based simulations, or treat it as a cheminformatics problem? How does this problem fit into the broader context of the world? For example, predicting a drug’s effect on a patient could be approached by simulating the entire person, which is currently impractical, or by modelling the effects statistically or causally. At some point, these different levels of models need to align, and domain scientists are crucial in mapping out this structured hierarchy of models. They help determine when assumptions are reasonable and when they are not. While ML tools cannot solve these problems independently, they can significantly aid in integrating different model components.

\section{How to tackle scientific problems? }

Armed with the above toolbox and perspectives, we then make recommendations on how to choose impactful problems in ML for chemistry and introduce a high-level structure of how ML problems are tackled.
We finally outline three areas for growth for research in ML for chemistry: breadth, depth, and scale.

\subsection{The Aspuru-Guzik/Whitesides rules for selecting important problems}

When one of us (Aspuru-Guzik) started the {\sl Matter Lab} then at Harvard University (2006-2018) and now at the University of Toronto (2018-), a set of rules for selecting significant problems began to emerge from intuition. In a hallway conversation with George Whitesides, who told Aspuru-Guzik he had similar guidelines, the three questions to ask before starting any research crystallized. We apply them at the {\sl Matter Lab} daily to select problems. Here, we specialize in ML in Chemistry, but these are widely applicable. The three questions emphasize novelty, importance, and feasibility in that order.

\subsubsection{Question 1: Has this problem been solved before?~~} 
Before starting a scientific endeavour, ask yourself this question. Of course, if it has not been solved before, your solution will be more impactful and lasting. Aim to be {\sl first } and not {\sl best}.

In the context of ML, improving on benchmarks, despite providing valuable signals of progress, is not the end goal of research.
This is particularly true in academic work, where research is not directly linked to profits and should be as novel as possible. Once new problems are established, the field will be opened to improve the results afterwards.

{\sl Will this work create a new connection between two areas?} When a paper introduces more questions than answers, the field grows. 
Simply applying an ML method to a new field can be novel. But novelty can be maximized if the proposed approach offers a new perspective, such as reframing a search problem as a generative modelling problem.

For example, we introduced 3D generative modelling to the field of rotational spectroscopy,\cite{cheng2024determining} which has opened the question of 3D structure elucidation from rotational spectroscopy alone. This is a clear example where {\sl first} beats any other research. There were no previous ML baselines to compare or benchmark our method to, because we introduced the first approach in the field!

\subsubsection{Question 2: Is what you set out to solve relevant to society?~~}
Before starting a scientific quest, consider whether it will help others widely. We, after all, operate in a domain of science that directly impacts human life. Humans and the entire biome interact with human-made chemicals every day. Think of problems that matter to the planet. Arguably, in the twenty-first century, which is riddled with environmental and political crises, this is quite relevant.\cite{aspuru2018matter}

Which audience will care?
What new tasks become within reach if this task is solved with significantly greater accuracy or speed?
For example, neural network potentials are significant because force fields are used in a large number of computational chemistry methods, which in turn predict properties and spectra. Solving this problem, therefore, touches a large audience.

Can the proposed method be tested experimentally if it solves a computational problem?
Approaches that can be experimentally validated have a much higher impact ceiling.\cite{watson2023novo,gorgulla2020open}
On the other hand, what is the worst-case scenario if the proposed approach ``doesn't work''?
If novelty is chosen carefully, this risk is mitigated because a method which solves an unbenchmarked problem is already state-of-the-art.

\subsubsection{Question 3: Is it remotely possible to attack this problem?~~}
Tackling something that is {\sl powerful}, yet {\sl within the reach of your resources} is key to success. The most effective and general publications will obviously have more impact. Therefore, aim for difficult and not low-hanging fruit work if what you wish is for your work to be remembered. 

In the context of ML, it would be useful to consider the following questions:
What are the available resources?
Is enough data available for the desired generalization performance?
Are there public code implementations?
Have similar problems been solved using the same framing?
For example, the success of 3D generative models in structure prediction on tasks such as conformer search and docking indicated that they can likely be successful in crystal structure prediction as well.

A crucial part of feasibility is controlling scope.
What is the minimal implementation of an algorithm that can solve this problem, yet have a broad impact?
How can success be evaluated within this problem scope?


\subsection{The structure of  data science and ML problems}

Machine learning and many data science problems have a general structure, as seen in many papers.
Once you begin on a chosen problem, the next considerations follow this hierarchy: (1) data, (2) problem framing, (3) method, and (4) evaluation. {\bf In our research group, we always think of these in order and in ranking.} For example, without {\sl data} a scientist will not be able to make progress. A publication that suggests a new method for old data will be less impactful than the publication that provided the data (and its ML application) in the first place. 

\subsubsection{What data are available?~~}
In machine learning, everything begins from the available data.
No method can be applied without it.
What is the size of the available data?
How easy is it to simulate new data?
What ground truth data are available, and what methods are available for validating a model’s predictions?
Anecdotally, when a dataset exceeds around 10,000 examples, generative models are more likely to generalize effectively.
Problems that are repeatedly solved in the community should be considered. Can these data be routinely recorded?
For instance, tasks like computing forces and conformer searches are standard in quantum chemistry, and the availability of this data has contributed to the success of neural force fields and 3D structure prediction. Additionally, data might not just be a static dataset but could include on-the-fly data acquisition, such as environments for agents or oracle functions for black-box optimization. It is because {\sl data is the ultimate resource} that our group embarked on the multi-year goal of developing and employing self-driving labs. We can eat our own dog food.


\subsubsection{What is a useful framing of the problem?~~}
The next critical task is to frame the problem usefully.
Framing is important not only to ensure selection of the right tools in Table \ref{tab:toolbox}, but also allows for benchmarking and theoretical analysis. Problem framing should be informed by domain knowledge: What specific challenges must be addressed to enable downstream tasks, such as experimental validation? 
For example, performing materials design by generating crystal structures as 3D unit cells may be difficult to translate into real materials, since experimentalists do not have atomistic control of structure.
Framing by itself can often determine the novelty and significance of the proposed research: Creating a new connection between a chemical problem and a ML problem generates novelty, and the potential step-function improvement in performance can improve significance.




Another way to approach problem framing is by asking how the data will be represented. Choosing a compact, information-rich, efficient-to-compute representation is a simple way to incorporate inductive bias and accelerate learning. However, as the bitter lesson shows, it is not essential to spend too much time on designing the ``perfect'' representation. Deep learning can automatically find ideal representations if the input representation contains all the necessary information and is available in large enough quantities.

\subsubsection{What model solves this problem?~~}

Once the problem is framed, the choice of model often becomes apparent and justified. What ML methods perform well for this task? 
Can simple methods solve this problem?
Established methods, such as Morgan fingerprints and XGBoost, remain strong baselines for property prediction,\cite{tom2023calibration} while genetic algorithms are strong baselines for molecular generation.\cite{tripp2023genetic}
If simple methods fail, are there new classes of algorithms suited for this problem? 
Is there existing code available online? 
It may be easier to first run the code before trying to understand the code.
How can a code implementation for solving another problem be modified as minimally as possible to solve the problem at hand?
Choose algorithms commensurate with the size and availability of data. With small datasets, classical machine learning still performs best.

This is perhaps the most critical paragraph of this publication: Golden advice to graduate students and postdocs, do not fall in love with the mermaids of new methodology. If older but proven methodology does the job, {\sl just use it!} Focus on the scientific contributions of your work. New methods should be developed when others truly have limitations. In other words, your new fancy super-duper autoencoder will not be as impactful in the long term as if you solve an essential chemistry or materials science question with an answer that lasts for ages.

\subsubsection{How will the proposed method be evaluated?~~}
Finally, the method must be evaluated according to reasonable metrics as informed by domain knowledge.
Do the metrics reflect the practical realities of downstream use cases of the proposed method?
For example, if you are generating and proposing new molecules, is it feasible for a chemist to synthesize them and test their properties?
Deciding appropriate metrics is vital because future work will likely adopt the same evaluation criteria.

\subsection{New problems: demanding impact from ML for chemistry}


Applying ML to chemistry can have a greater impact in terms of breadth of application, depth of consideration, and scale of execution.
In breadth, many more chemical problems can be formulated as ML problems and introduced to the ML community.
In depth, proposed methods can make stronger theoretical connections between both machine learning and computational chemistry, motivating further method development in each field.
Finally, at scale, ML for chemistry can aim at more significant problems requiring more data.
As concerns mount about reaching the limits of internet-scale data in language and vision, chemistry stands out as a situation where more data can be ``purchased'' through computational simulation or high-throughput experimentation.

\subsubsection{Solving problems in breadth.~~}

While in Sec \ref{SEC:taxonomy} we have witnessed the diversity of chemical problems that ML has been applied to, many areas of chemistry remain underexplored. 
In no particular order, we list a number of chemistry fields in which ML is still emerging: photochemistry,\cite{westermayr2020machine, axelrod2022excited} chemical education,\cite{du2024large} nuclear chemistry,\cite{morgan2022machine} agrochemistry,\cite{djoumbou2023cheminformatics} analytical chemistry,\cite{barone2021computational} electrochemistry,\cite{leong2024automated} astrochemistry,\cite{fried2024rotational} amorphous materials,\cite{zheng2024ab} soft materials,\cite{wang2024accessing} open quantum systems,\cite{ullah2024physics} environmental chemistry,\cite{zhu2023improved} and atmospheric chemistry, \cite{zhao2024leveraging} just to cite a few.
Within each field lie a number of tasks that could be formulated as ML problems, depending on the data available.
Tasks can also go beyond the idealization of pure, small organic molecules.
Heterogeneous materials, quantum materials, and complex mixtures present challenges that could particularly benefit from ML innovations. As mentioned in Sec. \ref{sec:analysis}, most substances in real-world situations are complex mixtures.

The key is not to ``force'' ML into these areas but to consider whether existing or novel tasks could be framed as ML problems listed in Table \ref{tab:toolbox}, facilitating iterative improvements and potentially leading to new algorithms.
In some situations, there is just not enough data to apply ML, but it remains that a simple way to guarantee novelty is to consider an underexplored field.

Coming back to our previous example, we are pretty happy to have applied ML to solve an essential structural determination in rotational spectroscopy: the first application of generative models to predict the 3D structure of molecules given their substitution coordinates.\cite{cheng2024determining} This is an example of a typical {\sl in-breadth} approach seeking multidisciplinary approaches and leaving our own comfort zone.

\subsubsection{Solving problems in depth.~~}

As we saw when discussing application-driven innovation in ML in Sec. \ref{sec:interdisciplinary}, chemical problems have motivated new algorithms and advanced ML theory.
Deep engagement with ML theory or theoretical chemistry generates novelty and significance and often leads to more robust empirical results.


Many ML methods such as graph neural networks and equivariant architectures were motivated or inspired by theoretical chemistry, and they are beginning to return to the favor.
Diffusion models were proposed in 2015, inspired by methods in statistical mechanics,\cite{sohl2015deep} and have since become state-of-the-art generative models enabling high-resolution text-to-image generation.\cite{ho2020denoising, song2020score, karras2022elucidating}
Nearly a decade later, new works have connected diffusion models to traditional tools in computational chemistry.
Diffusion models can simultaneously learn both coarse-grained force fields and a generative model,\cite{arts2023two} and can also be leveraged as a means for sampling and computing free energies.\cite{mate2024neural}
These works would not have been possible without deeper consideration of how diffusion models relate to free energy, or of the connection between diffused distributions and the ideal gas.

Furthermore, flow matching approaches derived from diffusion models relax the constraint of noising a data distribution to a pure Gaussian distribution and can instead connect two different distributions.
This has enabled learning of trajectories,\cite{neklyudov2023action,neklyudov2024wasserstein} which is beginning to be applied for transition path sampling of reactions.\cite{du2024doob}
These works create theoretical connections that may enable more techniques to transfer from computational chemistry to machine learning and vice versa.

In addition, whereas neural network potentials treat energy computation as a black-box function to be memorized, Hamiltonian prediction\cite{unke2021se} opens the box of Hartree-Fock theory, enabling access to the wavefunction, as well as a new tradeoff between accuracy and speed.
Self-consistency training\cite{zhang2024self} engages with this theory by removing the requirement of providing Hamiltonian matrices as labels, which has improved the speed of DFT overall.

Aiming for a concrete {\sl design} goal in collaboration with experimentalists also provides much-needed depth. Real-world problems often require the integration of ML with experimental data, and such collaborations can lead to breakthroughs that would not be possible in isolation. Large-scale collaborations between experts in quantum chemistry, machine learning, and organic materials chemistry enabled the discovery of new OLEDs.\cite{gomez2016design} In that work, we were among the first to demonstrate that fingerprint-based ML methods, intelligent screening methodologies, and experimental verification could lead to novel materials in a closed-loop philosophy.

Our group, more recently, spent five years in an international collaboration involving six research groups, which led to a delocalized, asynchronous closed-loop design that led to the best organic laser material to date (to our knowledge).\cite{strieth2024delocalized} In parallel, another multidisciplinary collaboration on closed-loop design\cite{angello2022closed} demonstrated that ML can teach us new chemical principles from these {\sl in-depth} materials science explorations.

\subsubsection{Solving problems at scale.~~}

The unreasonable effectiveness of scale, as shown by the bitter lesson (Sec. \ref{sec:bitter}), provides optimism for solving much more difficult problems.
Notorious problems like protein structure prediction were finally cracked by leveraging the scale of the Protein Data Bank.\cite{wwpdb2019protein, jumper2021highly}
Fast and quantum mechanically accurate atomic dynamics are being enabled by foundation force fields.\cite{batatia2023foundation,merchant2023scaling,yang2024mattersim}

For chemical problems which are already formalized in ML, progress can be accelerated just by increasing the scale of data and compute of these approaches.
Projects like the Open Catalyst Project demonstrate the potential of ML to drive large-scale advancements in chemistry.
By purchasing new data through computation and simulation and by designing better sampling algorithms, we can improve the rate of data generation, and take aim at scale. LLM agents, for example, could execute computational simulations to generate new training data, further accelerating research.

While training foundation models is often cited as a source of significant emissions, we should also be aware of the potential for compute to \emph{reduce} emissions.\cite{schilter2024balancing}
Better models could reduce the number of wet-lab experiments needed, or help design greener alternatives to current and future chemical processes, observing that the chemical industry makes up a large chunk of global emissions. 



\textit{Chemical space may be small.}
The often-cited estimated size of chemical space as $10^{60}$ fascinates us. But from a machine learning perspective, this space may be considered small. If we only consider black-and-white 28x28 images, the domain of the standard MNIST dataset of handwritten digits,\cite{deng2012mnist} this already has a size of $2^{28 \times 28} \approx 10^{236}$. Of course, the space of images is far sparser, given that the number of colour images in existence is 14.3 trillion $\approx 10^{13}$ images.\cite{broz2024many}
This is what makes deep learning impressive -- its ability to find structure within enormously high-dimensional spaces, just from showing a bunch of examples.
In the context of language, $10^{60}$ is just the number of 10-word sentences restricted to a vocabulary of 60 words, or the number of 10-sentence paragraphs restricted to 60 possible sentences. Natural language is evidently much larger.


Could these powerful capabilities be enough to turn theoretical musings into reality?
Imagine being able to atomistically simulate a cell on a macroscopic timescale, or to accurately model the effectiveness and stability of soft organic devices over years of use, or to discover new reactions ab initio. These are challenges that, until recently, seemed impossibly far beyond reach.
We are impressed that nanosecond simulation of an all-atom HIV capsid at DFT accuracy is possible with neural force fields.\cite{kozinsky2023scaling}
If modern image generative models can generate high-quality images at 1024x1024 resolution and higher,\cite{baldridge2024imagen} then what really stands in the way of simulating an entire cell at biological timescales?
If it is data, we are fortunate to have access to more and more complex simulations and self-driving labs which can generate high-quality data independently.
If the barrier is computing power, we are lucky enough to utilize the massive increases in computing power driven by mainstream AI.
If it is methods or experiments, then here is the call for action to all of us, multidisciplinary theoretical chemists of the twenty-first century: Let's transform our discipline together!








\section*{Acknowledgements}
This research was undertaken thanks in part to funding provided to the University of Toronto’s Acceleration Consortium from the Canada First Research Excellence Fund CFREF-2022-00042. 
A.A.-G. thanks Anders G.~Frøseth for his generous support. A.A.-G also acknowledges the generous support of the Canada 150 Research Chairs program. A.A. gratefully acknowledges King Abdullah University of Science and Technology (KAUST) for the KAUST Ibn Rushd Postdoctoral Fellowship.




\footnotesize{
\bibliography{corrections} 
\bibliographystyle{rsc}
}

\end{document}